\documentclass[3p, 12pt, times]{elsarticle}

\usepackage{graphicx}
\usepackage{caption}
\usepackage[labelfont=bf]{caption}
\usepackage{subcaption}

\usepackage{float}
\usepackage{amsmath}
\usepackage{amssymb}

\usepackage{algorithm}
\usepackage{algorithmicx}
\usepackage[noend]{algpseudocode}

\let\algIf\If
\let\algElse\Else
\usepackage{zed-csp}
\let\If\algIf
\let\Else\algElse

\usepackage{pifont}
\newcommand{\cmark}{\ding{51}}
\newcommand{\xmark}{\ding{55}}

\usepackage{longtable}
\usepackage{adjustbox}
\usepackage{lipsum} 
\usepackage{makecell}

\begin{document}

\begin{frontmatter}
\title {Towards Self-organized Large-Scale Shape Formation: A Cognitive Agent-Based Computing Approach}
\date{\vspace{-5ex}}
\author[add1]{Yasir R. Darr} 
\author[add1]{Muaz A. Niazi$^*$}
\address[add1]{Computer Science Department,\\ COMSATS Institute of Information Technology,\\ Islamabad, Pakistan \\
$^*$Corresponding author}


\begin{abstract}
Swarm robotic systems are currently being used to address many real-world problems. One interesting application of swarm robotics is the self-organized formation of structures and shapes. Some of the key challenges in the swarm robotic systems include swarm size constraint, random motion, coordination among robots, localization, and adaptability in a decentralized environment. Rubenstein et al. presented a system (``Programmable self-assembly in a thousand-robot swarm", Science, 2014) for thousand-robot swarm able to form only solid shapes with the robots in aggregated form by applying the collective behavior algorithm. Even though agent-based approaches have been presented in various studies for self-organized formation, however these studies lack agent-based modeling (ABM) approach along with the constraints in term of structure complexity and heterogeneity in large swarms with dynamic localization. The cognitive agent-based computing (CABC) approach is capable of modeling such self-organization based multi-agents systems (MAS). In this paper, we develop a simulation model using ABM under CABC approach for self-organized shape formation in swarm robots. We propose a shape formation algorithm for validating our model and perform simulation-based experiments for six different shapes including hole-based shapes. We also demonstrate the formal specification for our model. The simulation result shows the robustness of the proposed approach having the emergent behavior of robots for the self-organized shape formation. The performance of the proposed approach is evaluated by robots convergence rate.          
\end{abstract}

\end{frontmatter}

\section{Introduction}
\label{sec:intro}
A swarm robotics based systems have received a significant level of interest from the research community. These systems are inspired by the biological swarm, having social and collective behavior for achieving the task much bigger than their individual working capacity as discussed by Oh et al. in \cite{oh2017bio}. The same collective phenomenon is being used in the robotics field for performing everyday tasks in which group of robots interacts with each other to achieve global behavior as mentioned by Schmickl and Crailsheim in \cite{schmickl2008trophallaxis}. In this regard, much focus is being paid to the formation of structures and shapes using the self-organization process in swarm robotics.\\
 
Self-organization in a swarm robotics for the formation of diverse structures and patterns in a sustainable assembly is a key challenge for the researchers. It is purely based on the coordination and cooperation between robots as presented by Haghighi and Cheah in \cite{haghighi2012multi}, to produce and maintain the formation of a specific shape without any centralized controller. The aggregation and segregation of agents \cite{kumar2010segregation} along with their adaptive behavior are usually involved in achieving complex symmetries through the organization process in swarm robotics.\\   

A number of challenges are faced in the swarm robotic system for the self-organized shape formation. These challenges include the heterogeneous behavior of robots in a swarm, localization schemes described by de\'Sa et al. in \cite{de2016distributed}, adaptive behavior of robots in a decentralized environment, coordination among robots, and controlling the self-organized formation as presented by Cheng et al. in \cite{cheng2005robust}. Numerous studies show various techniques for addressing mentioned challenges in the context of implementation and control of the self-organized formation process in swarm robotics system.\\ 

The self-assembly based system proposed for the formation of shapes by Rubenstein et al. in \cite{rubenstein2014programmable}, in which experiment carried out for the formation process by applying the collective behavior algorithm. However, system is limited only to solid shapes with no holes, thus unable to make complex and diverse symmetries. The robots placed in an aggregated form having no dynamic localization and they move sequentially with no randomness. These issues are reported by Niazi in \cite{niazi2017technical}.

Another work regarding self-organization of agents presented by Copenhagen et al. for the group of robots in \cite{copenhagen2016self}. In this work, agents in a swarm made clusters through aggregation and segregation based on their alignments. While the proposed approach is dependent on the number of agents in a swarm and formed cluster contain both aligned and non-aligned agents in a two-dimensional space. In table \ref{tbl:problembStatment}, we analyze the mentioned systems regarding technique used, robustness of the system, and emergent behavior of swarm for the self-organization process.

\begin{table}[h]
\centering
\caption{Analysis for self-organization process achieved by the swarm robotics systems presented in \cite{rubenstein2014programmable} and \cite{copenhagen2016self}.}
\label{tbl:problembStatment}
\begin{adjustbox}{width=0.8\textwidth}

\begin{tabular}{c c c c c c c c}
\hline
\textbf{Ref.} & \textbf{ABM} & \textbf{\makecell{Swarm\\ limit}}  & \textbf{\makecell{Dynamic \\ localization}}& \textbf{\makecell{Random\\ motion}}  & \textbf{\makecell{Complex\\ shapes}} & \textbf{\makecell{Obstacle-\\avoidance}} & \textbf{\makecell{Heterogeneous\\ swarm}}\\ \hline

 \\ \cite{rubenstein2014programmable} & \xmark & \cmark & \xmark & \xmark & \xmark & \xmark & \cmark\\ \hline
 
\\ \cite{copenhagen2016self} & \xmark & \cmark & \cmark & \cmark & \xmark & \cmark & \cmark\\ 

\hline

\end{tabular}
\end{adjustbox}
\end{table}

Most of the issues presented in the previous work regarding self-organized shape formation can be resolved by developing ABM for MAS in the simulation-based environment. MAS is much helpful for observing behavioral variability in swarm robotic system for the formation of structures and shapes. The complex model-based systems developed in previous study using CABC in MAS \cite{2012cognitive}, 
can portray the self-organization process of swarm robots. These systems are capable of making robust and sustainable self-organization having the dynamic and adaptive behavior of robots in a swarm for the complex structures. \\
%

Our main contributions are as follows:
\begin{enumerate}
\item
	We develop a simulation model for the self-organized shape formation task of a swarm robotic system using ABM under CABC approach.
\item	
	We propose a self-organized shape formation algorithm to validate our model.
\item	
	We also demonstrate the formal specifications of our model.
 \end{enumerate}
The remaining structure of the paper is organized as follows: The section 2 shows the background and basic knowledge. In section 3, we include our research methodology for this study. The section 4 presents results and discussion for simulation-based experiments along with previous studies and their comparison. A conclusion and the future work are described in the section 5.  

\section{Background}
\label{sec:backgroud}
In this section, we discuss some of the concepts like swarm intelligence, swarm robotics using swarm intelligence, self-organization in swarm robots, ABM, and CABC.
\subsection{Swarm intelligence}\label{sec:SI}
Swarm intelligence refers to the collective behavior of decentralized artificial system having a bio-inspired approach as mentioned in \cite{karaboga2014comprehensive} by Karaboga et al. It is the soft form of a biological swarm exists in nature. Besides the structure and capabilities of the individual entity in a swarm, the global behavior is achieved through the social interactions among local individuals. It is not essential that a unique agent is useless but much simple as compared to the resultant intelligence or outcome behavior. The swarm intelligence allows the system to get the complete benefit from the swarm without having any sort of centralized control as discussed by Tan and Zheng in \cite{tan2013research}.  
\subsection{Swarm robotics}\label{sec:SR}
A swarm robotics field follows the swarm intelligence approach in which multiple robots interact with each other to achieve collective  behavior. The robots in a swarm have autonomous behavior and they act in a cooperative manner to achieve tremendously complex tasks as compared to their physical structure and capacity as suggested by Barca and Sekercioglu in \cite{barca2013swarm}. A design for coordination of the heterogeneous robots swarm is presented by Nishikawa et al. in \cite{nishikawa2016coordination}. The swarm robotics based systems are being used to perform several tasks such as partitioning of work, aggregation, pattern formation \cite{varghese2009review}, object sorting, navigation or path planning \cite{wu2016novel}, localization, and collective manipulation. These sort of tasks  are also given in a comprehensive study by Bayindir in \cite{bayindir2016review}.
\subsection{Self-organization}\label{sec:SO}
Self-organization process is used to attain overall assembly, which comes after local interaction among initially disordered components of the system \cite{niazi2008self}. It is a spontaneous process and is not organized by any agent externally. The resultant organization is completely decentralized and spread over all the components of the system, thus making it sustainable and robust \cite{zhang2014swarm}. The strategies for the self-organization are discussed by O\'Grady et al. in \cite{o2010self} for the group of mobile robots.\\

The process of organization for the swarm intelligence based swarm robotic system is being used for several purposes including the formation of structures. Such self-organized formation includes the aggregation or segregation of swarm robots by using odometry \cite{vardy2016aggregation} by Vardy and cue-based aggregation by Arvin and Turgut in \cite{arvin2016investigation}. The self-organized path formation discussed in \cite{sperati2011self} by Sperati et al. and by Nouyan et al. in \cite{nouyan2008path}. While producing patterns through a group of robots mentioned in the review by Bahceci et al. in \cite{bahceci2003review}.        
\subsection{Agent-based modeling}\label{sec:abm}
An ABM involves actions, characteristics of distinct agents and their interactions for describing real-world scenarios or any complex systems. An agent is a unique entity of the simulation having autonomous behavior, properties, and states. Agent-based representations using ABM can more easily be understood than mathematical models, it is because ABM created by simple rules regarding change in behavior and individual entities of the system. An ABM can also be used to represent the self-organization process by making a computational model of the MAS \cite{niazi2011novel}. Modeling of the self-organization in complex systems can be possible by using flexible, generic and effective ABM tools \cite{niazi2009agent}. 

\subsection{Cognitive agent-based computing}\label{sec:cabc}
The CABC framework is capable of making complex computational models based on the autonomous behavior of multi-agents. This involves the interaction and cooperation among agents based on learning techniques so to relate real-world complex systems. This was proposed and presented by Niazi in \cite{niazi2011towards}. It is capable of validating large-scale MAS by developing ABM before developing the actual system. It follows the different levels of modeling approaches such as descriptive or exploratory ABM, model validation, and complex network model. Laghari and Niazi used this framework in \cite{laghari2016modeling} for modeling the self-organizing communication network for the Internet of Things (IoTs).

\section{Methodology}
\label{sec:methodology}
\begin{figure}[H]
\begin{center}
\includegraphics[width = 8.0 cm, height = 8. cm]{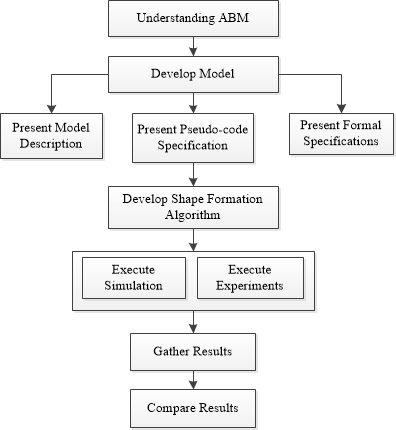}
\caption{Method block diagram for our research.}
\label{fig:method}
\end{center}
\end{figure}

In this section, we discuss the approach used in this study regarding development of ABM under CABC approach for self-organized shape formation. It starts from the understanding of ABM and development of the simulation model. To provide informal and formal specification of the model along with the pseudo-code specifications for the proposed shape formation algorithm. After that, applying of the algorithm on six different shapes to perform simulation-based experiments. At the end, analyze the results obtained from the simulation experiments. Figure \ref{fig:method} shows method block diagram of our research.
\subsection{Model description}
\label{sec:model-desc}
We present a model using ABM for self-organization of swarm robots at large-scale for shape formation task. In this model, agents act as a small programmable mobile robot initially placed at random locations in “NetLogo” environment before starting the self-organization process. These agents have heterogeneous behavior as very few are specially programmed to act as a seed robots. All other agents have the same behavior to follow the process and achieve collective behavior. The seed robots remain stationary throughout the organization process while other robots move randomly inside and outside shape coordinates. There is a deterministic type of environment for all the agents and predefined information given to all the agents in a swarm, about the shape to be formed in the coordinate system.\\  

Agents interact with each other as well as with the environment while moving randomly. The local interaction takes place among neighboring agents. The states of the agents are ``stationary", ``localize", and ``un-localize", that changes during the motion. Agents while moving enter the desired shape area and approach toward the seeds, or already localized robot to become stationary and localize. The formation process continues in this way until all the moving agents become stationary and localize in the desired shape area to form the shape completely.\\ 

There are three cases of termination of the formation process. Firstly, if the shape is complete, i.e. all the agents filled the shape and became stationary and localize. While there are no moving robots left, the process stops. Secondly, the shape is incomplete if there are few robots to fill the shape and no more robots present outside the shape area, the process stops again. Lastly, if the shape is formed completely while some robots still move outside the shape wait to join the formation process. These robots stop and separate themselves from the self-organization and the process stops. The resultant organization is completely distributed which is dependent upon all the agents as well as is sustainable and robust. \\
 
\subsubsection{States of formation process}
\label{sec:states}
\begin{figure}[H]
\begin{center}
\includegraphics[width = 8.0 cm, height = 7.0 cm]{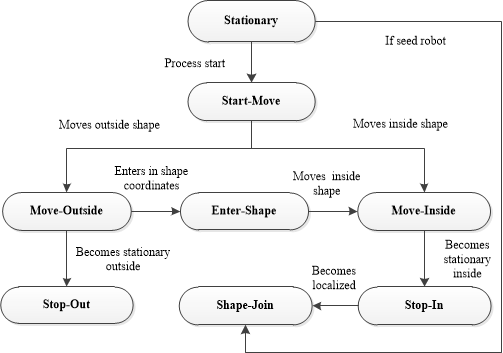}
\caption{State transition diagram for shape formation algorithm.}
\label{fig:method-states}
\end{center}
\end{figure}

The proposed shape formation algorithm comprises of different states of the robots during the organization process.  Before the process start, the system is in the initial state and robots are in the idle or stationary state. At the start of the process, all the robots move randomly outside or inside shape coordinates. Only four seed robots placed inside shape are stationary and localized throughout the process. A robot with the move-inside state changes its state to stop-in and joins the shape. Move-outside state changes to the move-inside state, when robot enters in the shape-area coordinates while moving randomly. The states transitions are shown in figure \ref{fig:method-states}.

 \subsubsection{Evaluation metric}
\label{sec:metric}
We consider number of parameters along with their values as evaluation metric in our proposed self-organized shape formation model. The ``region" parameter defines the dimension of the environment world in simulation tool. A ``number of robots" parameter is the dynamic user choice for the six different shapes, the given values are associated with one of the shape respectively. The communication or sense range is given in ``vision-radius" parameter having three values. A ``shapes" parameter consists of all the desired shapes. The ``performance evaluation" parameter has the average convergence rate of robots for the shape formation. A ``number of runs" parameter is for the experiment iterations. Table \ref{tbl:metric} shows all these parameters along with there values.   

\begin{table}[H]
\centering
\caption{Evaluation metrics of proposed approach for the self-organized shape formation.}
\label{tbl:metric}
\begin{tabular}{l | l }
\hline
\textbf{Parameter} & \textbf{Value}\\ \hline
\textit{Region} & 32 x 32 \\ 
\textit{No. of robots }& 1036, 1252, 566, 438, 1382,1040  \\ 
\textit{Vision-radius} & 1, 2, 3  \\ 
\textit{Shapes} & Starfish, k-letter, wrench, rectangle, tyre, spinner  \\  
\textit{Performance measure} & Average convergence rate  \\ 
\textit{No. of runs} & 4 (1, 10, 20,30 runs)   \\ 
\hline

\end{tabular}
\end{table}       

\subsection{Pseudo-code specification}
\label{sec:pcode-spec}

In this section, we present the pseudo-code for the proposed self-organized shape formation algorithm along with the detailed descriptions for each of the entity and behavior involved in the algorithm. We demonstrate the specifications for the agent involve in a process, all the global variables used, and all the procedures a proposed algorithm comprises of.     
  
\subsubsection{Agent design}

There is only one agent as ``Robot" used in our simulation model. A Robot act as an agent in a large-scale swarm having characteristics and states, performs various actions in order to model self-organized shape formation task.
\begin{algorithm}[H]
\caption*{Breed \textbf{Robot:} The robot agent used to represent robots in a swarm}
\emph{Internal Variables:} $<seed?, Stationary?, localize?, position$-$inside?>$
\begin{algorithmic}[1]
	 \Statex \textbf{seed?:} Specialize robots as a reference for all other robot 
	\Statex \textbf{stationary?:} All robots which are stationary
	\Statex \textbf{localize?:} All robots which are localize 
	\Statex \textbf{position-inside?:} All robots which are inside desired shape area
\end{algorithmic}
\label{algo:breed-robot}
\end{algorithm}

We first stated the agent used in the model environment to represent the robots in a swarm. Then we listed four internal variables for robots as ``seed?", ``stationary?", ``localize?", and ``position-inside?". Seeds are specialized robots which remain stationary and localize throughout the self-organization process. Robots which are static and not moving have stationary state, while localize state for robots placed at the correct position and position-inside state means robots which reside inside desired shape area whether moving or static.

\subsubsection{Global variables}
In this part, we describe global variable for our simulation. Input global variables are used during the initialization of the simulation environment through setup procedure.

\begin{algorithm}[H]
\caption*{\textbf{Global input:}$<num$-$robots, shape$-$type, vision$-$radius>$ }
\begin{algorithmic}[1]

\Statex \emph{Slider:}
 \Statex \hspace{\algorithmicindent} \textbf{num-robots:} Select number of robots in swarm 
	\Statex \hspace{\algorithmicindent} \textbf{vision-radius:} Communication range for robots interaction    
		
\Statex \emph{Chooser:}
	\Statex \hspace{\algorithmicindent} \textbf{Shape-type:} Select desire shape for the formation
\label{algo:Input-globals}
\end{algorithmic}
\end{algorithm}

Three global input variables used in our simulation model. Two variables ``num-robots" and ``vision-radius" are used through the slider, which is GUI element of NetLogo. Number of robots available at the start of the self-organization process are selected by ``num-robots". The ``vision-radius" used for choosing a radius for communication among neighboring robots. The ``shape-type" input variable provided by the chooser, which is also a GUI element for a simulation model of NetLogo. The ``shape-type" has a list of six desired shapes to be formed by robots. The list includes star, wrench, k-letter, rectangle, tyre, and spinner shape choices.

\subsubsection{Procedures}
We present all used procedures and their specifications in this section. 

\begin{enumerate}
\item
\textit{Setup}

\begin{algorithm}[H]
\caption*{Procedure \textbf{setup:} Creating simulation environment}
\emph{Input:} Global variables\\
\emph{Output:} Initialize the whole environment\\	 
\textbf{begin}
\begin{algorithmic}[1]

	\State Clear all
    \State Set-patches
    \State Set-shape
    \State Set-robots
  
\end{algorithmic}
\textbf{End}
\label{algo:pro-setup}
\end{algorithm}

The specification model for setup procedure is presented, which is used to create and initialize the main environment of the simulation model by considering all global variables. It clears all the previous states and stuffs present in the environment. After this, ``set-patches", ``set-shape", and ``set-robots" procedures are called which we discuss next.\\     
\item
\textit{set-patches}

\begin{algorithm}[H]
\caption*{Procedure \textbf{set-patches:} Creates number of patches in an environment world of NetLogo}
\emph{Input:} World size for x and y coordinate\\
\emph{Output:} Generates patches with boundary wall\\	 
\textbf{begin}
\begin{algorithmic}[1]

	\ForAll {patches} 
	   \State set color $=$ white
	     \If {neighboring patches $<$ 8}
	         \State set color $=$ black
	     \EndIf
		\EndFor
  
\end{algorithmic}
\textbf{End}
\label{algo:pro-set-patches}
\end{algorithm}

In the set-patches procedure, patches are created in the world as specified in the environment setting, which is a basic element in NetLogo. The x,y coordinate values is set along with the patch size for the number of patches. The color of all patches set to white, while black color for the corner patches to consider them as a closed boundary wall for robots. \\

\item
\textit{set-shape}

\begin{algorithm}[H]
\caption*{Procedure \textbf{set-shape:} Import selected shape on the patches of the world}
\emph{Input:} List of shapes in the chooser menu ``shape-type"\\
\emph{Output:} Generates selected shape information to the model\\	 
\textbf{begin}
\begin{algorithmic}[1]
	 
	     \If {shape-type $=$ ``star" }
	     \State import-star-shape
	     \EndIf
		\If {shape-type $=$ ``wrench" }
	     \State import-wrench-shape
	     \EndIf\If {shape-type $=$ ``k-letter" }
	     \State import-k-letter-shape
	     \EndIf\If {shape-type $=$ ``rectangle" }
	     \State import-rectangle-shape
	     \EndIf
	     \If {shape-type $=$ ``tyre" }
	     \State import-tyre-shape
	     \EndIf
	     \If {shape-type $=$ ``spinner" }
	     \State import-spinner-shape
	     \EndIf
  
\end{algorithmic}
\textbf{End}
\label{algo:pro-set-shape}
\end{algorithm}

In this procedure, the shape is selected by the global variable `shape-type". The selected shape is imported on the patches and its information is given to all robots.\\   

\item
\textit{set-robot}

\begin{algorithm}[H]
\caption*{Procedure \textbf{set-robot:} To create and set robots}
\emph{Input:} Number of robots from ``num-robots" slider\\
\emph{Output:} Generates number of robots at random locations\\	 
\textbf{begin}
\begin{algorithmic}[1]

	     \If {count robots = 0 }
	     \State create num-robots
	     \EndIf
		\If {robot-id $>=$ 0 and $<=$ 3 }
	   		\State {set shape = ``circle" }
	   		\State set color = ``red"
	   		\State set seed? = true
	   		\State set xy-coordinates near origin
	   		\State set stationary? = true
	   		\State set localize? = true  
	    
	    \Else
	    	\State set shape = ``directional-circle"
	   		\State set color = ``magenta"
	   		\State set xy-coordinates  random 
	   		\State set seed? false
	  		\State set stationary? false
	  		\State set localize? false 
		\EndIf
  		\ForAll {robots}
  		\If {xy coordinate inside shape}
  		\State set position-inside = true
  		\Else
  		\State set position-inside = false
  		\EndIf
  		\EndFor
\end{algorithmic}
\textbf{End}
\label{algo:pro-set-robots}
\end{algorithm}
 
In the set-robots procedure, existing number of robots are checked first, then number of robots selected by the ``num-robots" slider are created. First four robots dedicated as seed robots. The shape for seeds set as ``circle", color as ``red" and location near the origin, while values for internal state variables ``seed?", ``stationary?", ``localize?", and ``position-inside?" as true. For all other robots, the shape is chosen as ``directional-circle", color as ``magenta" and random coordinates as location, while values for internal state variables are set false. The locations of robots in the environment world are checked to set ``position-inside?" value as true or false.\\     

\item
\textit{Procedure go}

\begin{algorithm}[H]
\caption*{Procedure \textbf{go:} Main shape-formation algorithm}
\emph{Input:} Number of robots from num-robots slider\\
\emph{Output:} Generates shapes by convergence of robots\\
\emph{Execution:} Called repeatedly during simulation\\	 
\textbf{begin}
\begin{algorithmic}[1]
	\While {stoping condition = true}
	 \ForAll {non-stationary robots}
	    \If{coordinates = outside-shape} 
			\State set position-inside? = False
			\State Move Random
			\If{patch at vision-radius = black}
			\State Rotate Random 360
			\Else 
			\State Move Random      
			\EndIf
			\If{any moving robot at vision-radius}
			\State Rotate Random 360
			\Else 
			\State Move Random      
			\EndIf
		\Else\space{(inside shape coordinates)}
			\State set position-inside? = True
			\State Move Random
			\If{any moving robot at vision-radius}
			\State Rotate Random 360
			\Else 
			\State Move Random      
			\EndIf
			\If{any empty-patches at neighbors of localize robots}		
			\State face empty-patches
			\State move-towards-location
			\State set stationary? = true
			\State set localize? = true
			\EndIf
		
		\EndIf		
		\EndFor  		
  		\EndWhile
\end{algorithmic}
\textbf{End}
\label{algo:pro-go}
\end{algorithm}
%
In this procedure, the main working for the formation process is shown. A number of robots randomly moving outside or inside shape coordinates in the cartesian coordinate system of the environment world. The ``position-inside" variable is set false outside shape and true for inside shape coordinates. Non-stationary robots avoid collision with each other in ``move-until-empty-ahead" procedure. While inside shape coordinates, non-stationary robots check and proceed towards empty neighbor patches of seeds or any stationary and localized non-seed robot in ``move-towards-location" procedure. After reaching to the desired location, robots set their values for ``stationary?" and ``localize?" as true. Figure \ref{fig:algo} shows the flow of the shape formation algorithm.     

\begin{figure}[H]
\begin{center}
\includegraphics[width = 10.0 cm, height = 7.0 cm]{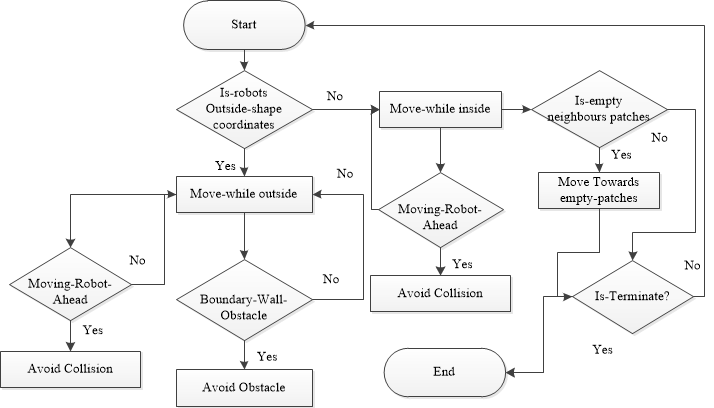}
\caption{Flowchart for the proposed shape formation algorithm.}\label{fig:algo}
\end{center}
\end{figure}

\item
\textit{boundary-wall}

\begin{algorithm}[H]
\caption*{Procedure \textbf{boundary-wall:} Avoiding boundary wall obstacle}
\emph{Input:} Moving robots and their locations  \\
\emph{Output:} Directions for moving robots\\	 
\emph{Execution:} Repeatedly called by ``go" procedure\\	
\textbf{begin}
\begin{algorithmic}[1]

	\ForAll {non-stationary robots} 
     	\If {outside-shape coordinates}
     		\If {next patch = black}
	         \State Rotate Random 360 
	         \Else
	         \State Move Random      
	     		\EndIf
	     \EndIf
		\EndFor
  
\end{algorithmic}
\textbf{End}
\label{algo:pro-wall}
\end{algorithm}
In this procedure, all the moving robots outside shape coordinates check next location in their direction for the boundary wall obstacle. When reach to the boundary, robots return and divert their direction without hitting the wall obstacle.\\    

\item
\textit{move-until-empty-ahead}

\begin{algorithm}[H]
\caption*{Procedure \textbf{move-until-empty-ahead:} Avoiding collision between moving robots}
\emph{Input:} Moving robots and their neighbors    \\
\emph{Output:} Movement and directions of moving robots\\	 
\emph{Execution:} Repeatedly called by ``go" procedure\\	
\textbf{begin}
\begin{algorithmic}[1]

	\ForAll {non-stationary robots} 
     	\If {any moving robot ahead}
     		 \State Rotate Random 360 
	         \Else
	         \State Move Random      
	  	     \EndIf
		\EndFor
  
\end{algorithmic}
\textbf{End}
\label{algo:pro-collision}
\end{algorithm}

All moving robots avoid the collision with other robots by checking any robot ahead in their direction of motion. If so, robots change their motion direction. \\

\item
\textit{move-towards-location}    

\begin{algorithm}[H]
\caption*{Procedure \textbf{move-towards-location:} Robots choose and move towards candidate location to converge}
\emph{Input:} Moving robots and their locations  \\
\emph{Output:} Robots stop adjacent to localized robot\\	 
\emph{Execution:} Repeatedly called by ``go" procedure\\	
\textbf{begin}
\begin{algorithmic}[1]

	\ForAll {non-stationary robots} 
     	\If {inside-shape coordinates}
     		 \If {any empty-patches at neighbors of localized}
	         \State face empty-patches
	         \State move-towards empty-patches
	         \State set stationary? = true
	         \State set localize? = true 
	     	\EndIf
	     \EndIf
		\EndFor
  
\end{algorithmic}
\textbf{End}
\label{algo:pro-stop}
\end{algorithm}

Inside the desired shape coordinates, moving robots find an empty patch among neighbor patches of the seeds or other localized robots. They proceed toward that location and set values for ``stationary?" and ``localize?" as true.       

\end{enumerate}

\subsubsection{Behavior experiment}
The behavior of the moving robots along with their states is observed in the plots and experiment specification model.

\begin{enumerate}
\item
\textit{Produce plot}
\begin{algorithm}[H]
\caption*{Procedure \textbf{produce-plot:} To plot current values of the simulation model  }
\emph{Input:} NA  \\
\emph{Output:} Generates plot during simulation run\\	 
\emph{Execution:} Generated by ``go" procedure\\	
\textbf{begin}
\begin{algorithmic}[1]
	\Statex Number of stationary robots
	\Statex Number of localize robots
	\Statex Robots with position-inside = true
	\Statex Number of empty-patches inside shape

\end{algorithmic}
\textbf{End}
\label{algo:pro-plot}
\end{algorithm}

The ``produce-plot" procedure generates plot during the execution of the simulation to monitor the convergence of robots for shapes formation. It monitors a number of robots along with the state variable values for ``stationary", ``localize", and ``position-inside" as true.\\

\item
\textit{Experiment}  

\end{enumerate}
\begin{algorithm}[H]
\caption*{\textbf{Experiment:} Exhibits the convergence behavior of robots regarding six shapes}
\begin{algorithmic}
\Statex \emph{Input:} $< num$-$robots, shape$-$type>$ 
\Statex \emph{Setup procedure:} $<setup>$	 
\Statex \emph{Go procedure:}  $<shape$-$formation>$	
\Statex \emph{Repetition:} 10\\
\hrulefill

	\Statex \emph{Input:} 
	\Statex \hspace{\algorithmicindent} \textbf{num-robots:}	1036, 											566, 1352,438,1282, 1040  
	\Statex \hspace{\algorithmicindent} \textbf{shape-type:}	star, 					wrench, k-letter, rectangle, tyre, spinner
	\Statex \emph{Report:}
		\Statex \hspace{\algorithmicindent} \textbf{stationary:}	Number of stationary robots
		\Statex \hspace{\algorithmicindent} \textbf{localize:}	Number of localized robots
		\Statex \hspace{\algorithmicindent} \textbf{empty-patches:}	Number of empty-patches
		\Statex \emph{Stop condition:}  Self-organization process stops, when all robots become stationary and localize to form the desired shape completely.
\end{algorithmic}
\label{algo:pro-exp}
\end{algorithm}

A simulation experiment used in our model for the six shapes convergence analysis by varying number of robots. The average convergence rate calculated for the ten repetitions along with the shape completeness in term of localize, unlocalize robots, and empty-patches in the formed shape.

\subsection{Formal specification}
\label{sec:formspec}

The formal specification is a technique used for modeling different types of systems mathematically. A system developer can comment about the completeness of the model for all aspects needed in specific abstraction level. The one significant feature of formal specification is that through modeling, it can effectively represent the real-world scenarios at various abstract levels. A formal specification language is known as ``Z" developed by Oxford University in 1970. It can easily be converted into executable program code \cite{woodcock1996using}. A specification using ``Z" comprises of sets and predicates from the mathematical notation \cite{bowen1996formal}. This specification language is being used for modeling large-scale real-world complex systems. It is also used to relate the agent logic in MAS \cite{d2004understanding}.

\subsubsection{A Formal specification for self-organized shape formation}

\begin{figure}[H]
\begin{subfigure} {0.5\textwidth}
\centering
\includegraphics[width = 5.0 cm, height = 5.0 cm]{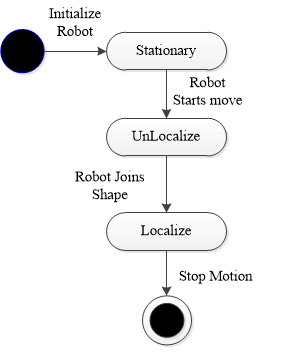}
\caption{}
\label{fig:RobotState}
\end{subfigure}
\begin{subfigure}{0.5\textwidth}
\centering
\includegraphics[width = 5.0 cm, height = 5.0 cm]{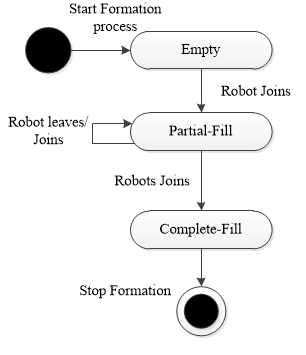}
\caption{}
\label{fig:ShapeState}
\end{subfigure}
\caption{The States for the robot and shapes \textbf{(a)} Robot states comprises of ``Stationary", ``UnLocalize", ``Localize", \textbf{(b)} Shape states comprises of ``Empty", ``Partial-Fill", ``Complete-Fill" .}
\label{fig:FormalStates}
\end{figure}

We present the formal descriptions of our self-organized shape formation model using z-language for formal specification. The purpose of this is to validate all the informal requirements of the system by using formal specification. In this study, robot and shape are two entities involved in the formation process. We use ``sets" along with operation or state ``schemas", for declaring user-defined entities such as robot and shape regarding shape formation task. The states of the robot and shape are shown in figure \ref{fig:FormalStates}. 
	
\begin{enumerate}
\item 
\textit{Sets and schemas\\}

 A ``ROBOT" is a set of all the robots in a swarm for the shape formation.

\begin{zed}
[ROBOT] 
\end{zed}

We defined the ``max-xcor" and ``min-xcor" coordinate values for x coordinate as 35 and -35 respectively and similarly for the other two coordinates y and z. While ``maxRobot" is user defined global input variable for defining maximum number of robots in a swarm.\\  

\begin{axdef}
maxRobot:\nat\\
xcor, ycor, zcor : \num
\where
maxRobot \geq 0\\
min$-$xcor=$-32$\\
max$-$xcor=$-32$\\
min$-$ycor=$-32$\\
max$-$ycor=$-32$\\
min$-$zcor=$-32$\\
max$-$zcor=$-32$
\end{axdef}

%


We use  a free type definition for the two types of robots in a swarm i.e. seed and non-Seed. A formal definition for ``RobotType" is:

\begin{zed}
RobotType ::=  seed | nonSeed
\end{zed}

Robot schema defines a robot placed randomly having  x,y,z coordinates in the cartesian coordinate system bounded between ``min" and ``max" values for coordinate parameters. A  variable of the type ``RobotType" introduced in this schema to define a robot as seed or non-seed.

\begin{schema}{Robot}
type:RobotType\\
x,y,z : \num
\where
x\geq min$-$xcor\\
x \leq max$-$xcor\\
y \geq min$-$ycor\\
y \leq max$-$ycor\\
z \geq min$-$zcor\\
z \leq max$-$zcor
\end{schema}

We use free type definition for states of the desired shape to be formed. Elements for this free type are ``empty", ``partial", and ``complete" for the empty shape, partially formed shape, and completely formed shape respectively.

\begin{zed}
ShapeState ::= empty | partial  | complete
\end{zed}

The schema describes the shape to be formed in coordinate system and shape dimension variable value obtained from the product of coordinates value of the shape. We define a variable ``sState" of the type ``ShapeState" who's value can be ``empty", ``partial", and ``complete".\\

All robots in a swarm carry characteristics like shape, color, state and position. The values for these features vary among robot, so we define the finite sets of robots  as ``stationary", ``unLocalize" and ``localize" for the states of robot.  

\begin{schema}{Shape}
p,q :\num\\
shapeDimention : \nat\\
sState:ShapeState\\
stationary: \finset ROBOT\\
unLocalize: \finset ROBOT\\
localize: \finset ROBOT\\
\where
p\geq min$-$xcor\\
p \leq max$-$xcor\\
q \geq min$-$ycor\\
q \leq max$-$ycor\\
shapeDimention = p*q\\
localize \leq maxRobot\\
localize \leq shapeDimension

\end{schema}

At the start of the formation process, shape is in empty state. That is,  still no robot has converged and joined the shape while ``localize" set is still a null set. 

\begin{schema}{InitShape}
Shape
\where
localize = \emptyset\\
sState = empty
\end{schema}

When process starts, all the non-seed robots in a swarm start moving randomly and change their state from ``stationary" to ``unLocalize". Robots are added into the finite set of ``unLocalize" and eliminated from ``stationary" set as shown in the schema ``StartMove"  given below.

\begin{schema}{StartMove}
\Delta Shape\\
Robot\\
robot? :ROBOT
\where
robot? \in stationary\\
type = nonSeed\\
robot? \notin unLocalize\\
unLocalize' = unLocalize \cup \{ robot? \} \\
stationary' = stationary \setminus \{ robot? \}
\end{schema}

Robots moving randomly enter and joins the shape upon reaching next to one of the localized seed robot near origin or other localized non-seed robot inside the shape. A robot belongs to ``unLocalize" set, becomes localize and added to ``localize" set as well as removed from ``unLocalize" robot set. The state of the shape changes from ``empty" to ``partial".  

\begin{schema}{JoinShape}
\Delta Shape\\
robot? :ROBOT
\where
\# localize < shapeDimension\\
robot? \in unLocalize\\
robot? \notin localize\\
localize' = localize \cup \{ robot? \} \\
unLocalize' = unLocalize \setminus \{ robot? \} \\
sState' = partial
\end{schema}

While formation process continues for the shape, procedure given in ``JoinShape" schema is repeated. When shape is completely formed by a number of localized robots, the ``sState" variable changes its value from ``partial" to ``complete". 

\begin{schema}{ShapeFormation}
\Delta Shape\\
robot?:ROBOT
\where
\# localize < shapeDimension\\
sState = partial\\
robot? \notin localize\\
localize' : localize \cup \{ robot? \} \\
unLocalize' : unLocalize \setminus \{ robot? \} \\
\# localize' = shapeDimension\\
sState' = completed
\end{schema}

We consider both perspectives for working of the  ``JoinShape", as successful and error outcome of the schema. The success outcome involves robots entering and joining the shape formation while for the error outcome, there will be two reasons for that. The shape is already formed so that no further robot can join the shape for formation. Secondly, if a total number of the robot is less than shape dimension value i.e. all robots belong to ``localize" set and shape is not completely formed. This scenario is presented in the table \ref{tbl:freeType}

\begin{table}[H]
\centering
\caption{Free type definition table for ``JoinShape" schema }
\label{tbl:freeType}
\begin{tabular}{| c | c | c |}\hline
\textbf{Schema} & \textbf{Pre-condition for success} &\textbf{Condition for error}\\ \hline  
JoinShape & \makecell{\# localize $<$ shapeDimension \\ 
 robot? $\notin$ localize} &\makecell{ 
Shape is alreadyfilled: \\
\# localize = ShapeDimension \\
robot? $\notin$ localize\\\\
Shape is incomplete: \\
maxRobot $<$  ShapeDimension \\
robot? $\in$ localize\\
} 
\\ \hline 
\end{tabular}
\label{tbl:shapeStatus}
\end{table}  

We define a free type definition ``REPORT" for success and error outcome for the shape formation process. The elements of this free type are ``success", ``alreadyComplete", and ``incomplete".    
 
\begin{zed}
REPORT ::= success | alreadyComplete | incomplete 
\end{zed}

\item
\textit{Schemas for Success and Error}\\

A robot joins the shape for the formation successfully in the ``JoinShape" schema. The report variable gives ``success" as output value.

\begin{schema}{Success}
report! : REPORT
\where
report!= success
\end{schema}

The shape is completed if the number of localized robots inside the desired shape is equal to the shape's dimension value. The report variable gives ``alreadyComplete" as outpu

\begin{schema}{AlreadyComplete}
\Xi Shape\\
report! : REPORT
\where
sState = complete\\
\# localize = shapeDimension\\
report!= alreadyComplete
\end{schema}

The shape is partially formed, if a number of localized robots is less than the shape's dimension value. The report output variable gives ``incomplete" as value. 

\begin{schema}{Incomplete}
\Xi Shape\\
report! : REPORT
\where
sState = partial\\
\# localize < shapeDimension\\
report!= incomplete
\end{schema}

\item 
\textit{Schema calculus\\}
We combine success outcome of ``JoinShape"schema for the shape formation process by using conjunction along with error outcome by using disjunction. 

\begin{zed}
(JoinShape \land success) \lor alreadyComplete \lor incomplete 
\end{zed}

\end{enumerate}

\section{Results and Discussion}
\label{sec:RisDis}

This study focused on self-organization of robots in a swarm at a large-scale by developing a simulation model using ABM under CABC for six different shapes. We performed a number of simulation experiments to analyze robots convergence. The proposed algorithm  is capable to form complex symmetries and diverse shapes including hole-based shapes. The self-organization process is robust without depending on the size of swarm i.e. group of robots, as remaining extra moving robots separate themselves from the process. While the selection and localization of seeds is based on the robot id among a variable number of robots. A randomness in motion and initial locations is used in our approach. The obstacle-avoidance mechanism used in term of closed boundary walls as well as collision-avoidance among randomly moving robots also addressed. We analyzed average convergence time for the completely formed shapes by the fixed number of robots by performing (n=10) simulation experiments along with 95\% confidence interval. The behavior analysis of our proposed approach shows that the average time convergence by the large group size robots is less than that for small group size robots.

\subsection{Environment overview}
\label{sec:resmodel}

\begin{figure}[H]
\begin{center}
\includegraphics[width = 12.0 cm, height = 5 cm]{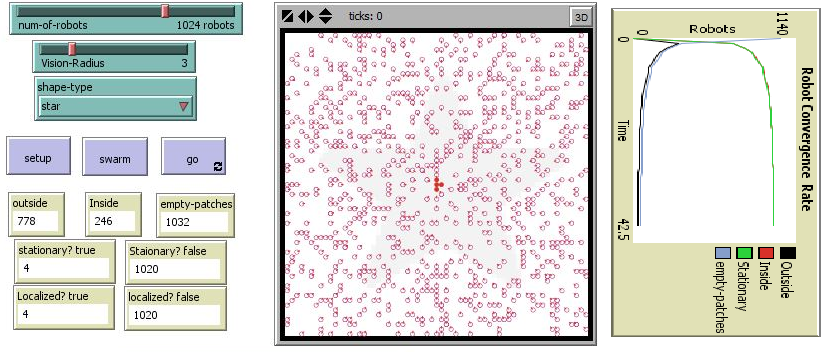}
\caption{A screen shot of the interface design for self-organized shape formation model in Netlogo.}
\label{fig:model}
\end{center}
\end{figure}

We used NetLogo as the simulation environment for our self-organized shape formation model. NetLogo is a tool used for the ABM at an extensive level by the number of researchers across the multiple domains \cite{wilensky1999netlogo}. It is based on the programming language ``Logo" used for the environment refers to ``World" in the graphical interface of the tool. Agents exist in the world named as a ``Turtles", which carry various characteristics and capabilities for performing a large number of actions. The state and behavior of agent changes against these actions performed in the world environment. Figure \ref{fig:model} shows the interface design of the NetLogo for our ABM based model.  

\subsection{Self-organized shape formation}
\label{sec:resconv}
\begin{figure}[H]
\begin{subfigure}{0.5\textwidth}
 \centering
   \includegraphics[width = 5 cm, height = 4 cm]{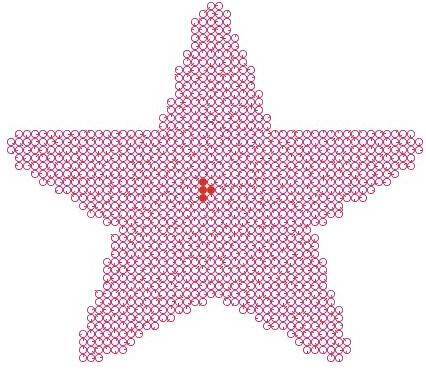}
  \caption{}
  \vspace*{1 cm}
\end{subfigure}%
\begin{subfigure}{0.5\textwidth}
 \centering
  \includegraphics[width = 5 cm, height = 4 cm]{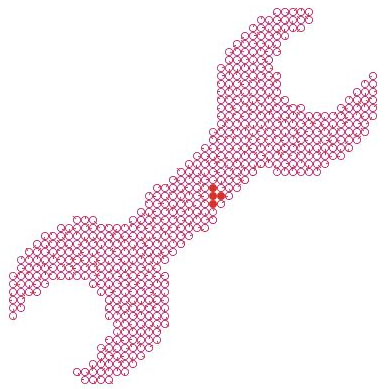}
  \caption{}
    \vspace*{1 cm}
\end{subfigure}
\begin{subfigure}{0.5\textwidth}
  \centering
  \includegraphics[width = 5 cm, height = 4 cm]{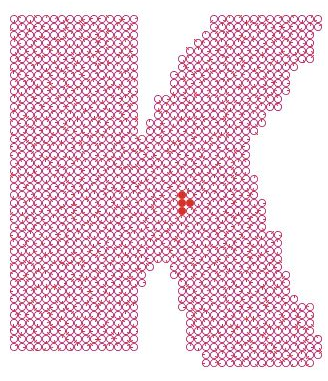}
  \caption{}
  \vspace*{1 cm}
\end{subfigure}
\begin{subfigure}{0.5\textwidth}
  \centering
  \includegraphics[width = 5 cm, height = 4 cm]{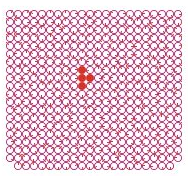}
  \caption{}
  \vspace*{1 cm}
\end{subfigure}
\begin{subfigure}{0.5\textwidth}
 \centering
  \includegraphics[width = 5 cm, height = 4 cm]{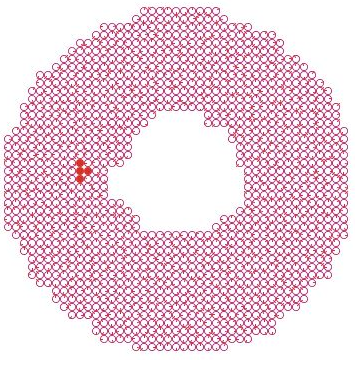}
  \caption{}
  \vspace*{1 cm}
\end{subfigure}
\begin{subfigure}{0.5\textwidth}
  \centering
  \includegraphics[width = 5 cm, height = 4 cm]{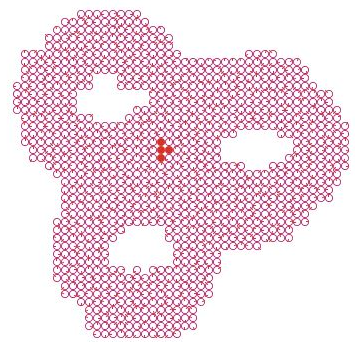}
  \caption{}
  \vspace*{1 cm}
  
\end{subfigure}
\caption{Desired six shapes formed by applying proposed shape formation algorithm in our simulation model.}
\label{fig:allShapes}
\end{figure}

We developed a simulation model for the formation of six shapes. These shapes include ``star", ``wrench", ``k-letter", ``rectangle", ``tyre", and ``spinner-shape". All final shapes formed by the self-organization of swarm robots in our simulation model by applying proposed shape formation algorithm developed in Netlogo are shown in figure \ref{fig:allShapes}.   

\subsection{Experiment results}
\label{sec:resexp}

\begin{figure}[H]
\begin{subfigure} {0.5\textwidth}
\includegraphics[width = 9.0 cm, height = 6.5 cm]{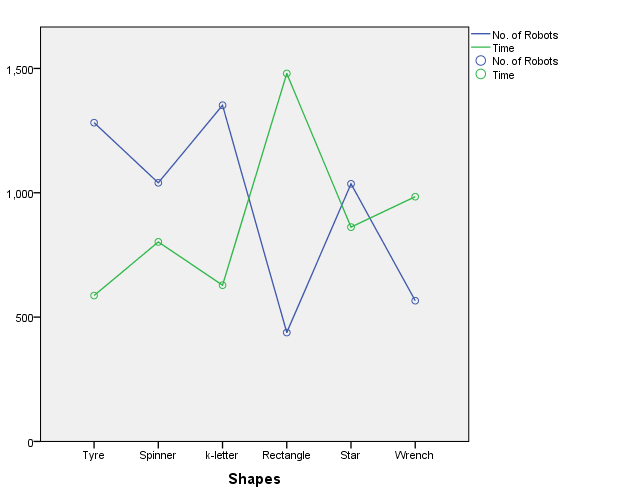}
\caption{}
\label{fig:shapesCon}
\end{subfigure}
\begin{subfigure}{0.5\textwidth}
\includegraphics[width = 7 cm, height = 6.5 cm]{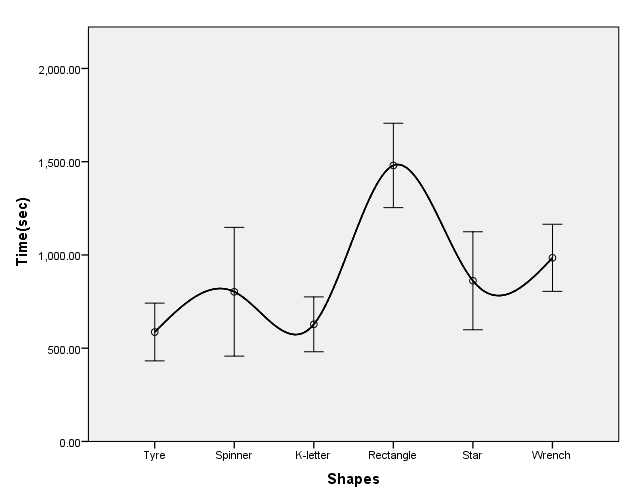}
\caption{}
\label{fig:shapesConI}
\end{subfigure}
\caption{Experiment results for all six shape showing convergence time for number of robots along with 95\% confidence interval.}
\label{fig:all-result}
\end{figure}

In this section, we discuss the simulation-based experiment's results obtained by applying proposed self-organized shape formation algorithm. The results include number of robots a final desired shape comprises, the average time taken to complete the formation of desired shape after running n=10 experiment for the number of robots, and the dimensions of the desired shape formed as shown in the table \ref{tbl:new-results}. The convergence rate i.e. the total number of robots involved in the formation process for the desired shape in a unit time along with the 95\% confidence interval for convergence time are shown in the figure \ref{fig:all-result}. \\

Next we describe the simulation results for all six shapes individually. We analyzed the behavior during simulation experiments by varying the number of robots and averaging their convergence time for the shape formation along with 95\% confidence interval value. A completeness of the shapes in term of a total number of localized, unlocalized robots, and empty spaces or patches within the formed shape is also being evaluated. In addition, we observed convergence variations among five different robots group size values for the specific shape at the completion of the simulation. While at the end of this section, we also discuss the  previously achieved results present in the study  \cite{rubenstein2014programmable}.   

\subsubsection{Star shape}
\label{sub-sec:star}
\begin{figure}[H]
\begin{subfigure}{0.5\textwidth}
  \includegraphics[width = 6.7 cm, height = 5 cm]{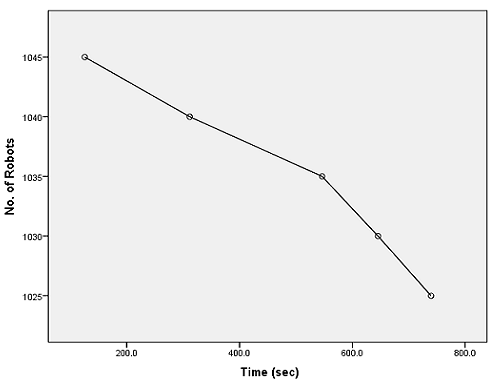}
  \caption{}
  \label{fig:starConTime}
  \vspace*{0.3 cm}
\end{subfigure}%
\begin{subfigure}{0.5\textwidth}
  \includegraphics[width =6.7 cm, height = 5 cm]{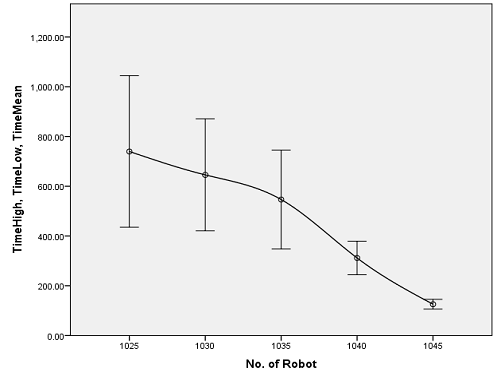}
  \caption{}
  \label{fig:starConI}
    \vspace*{0.3 cm}
\end{subfigure}\\
\begin{subfigure}{0.5\textwidth}
  \includegraphics[width=8.5 cm, height = 5 cm]{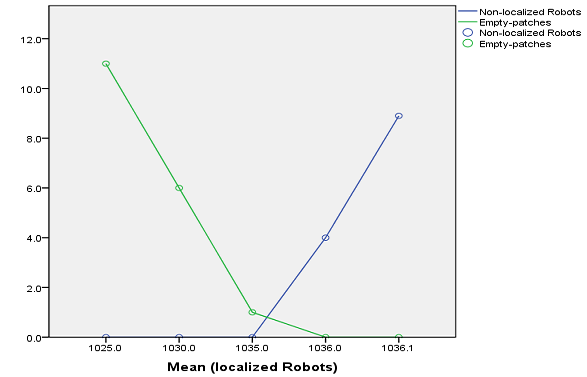}
  \caption{}
  \label{fig:starShapCom}
\end{subfigure}
\begin{subfigure}{0.5\textwidth}
  \includegraphics[width=8.5 cm, height = 5 cm]{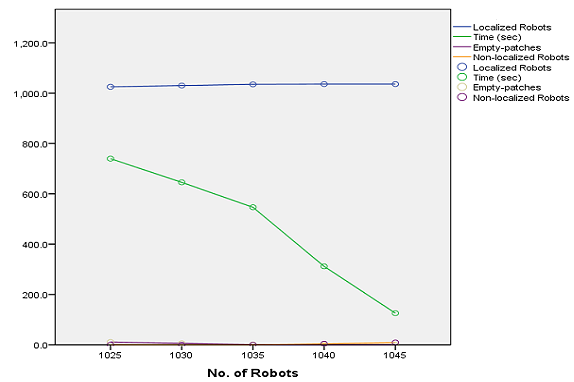}
  \caption{}
  \label{fig:starConVar}
\end{subfigure}

\caption{Experiment results for the ``star" shape, show the time convergence and shape completeness for five different robot groups.}
\label{fig:star}
\end{figure}

By applying the proposed algorithm for a number of robots in our simulation model, robots converge to form a predefined ``star" shape as imported into the model and shape information given to all agents at the start of the simulation process. We considered number of robots in five groups as (1025, 1030, 1035, 1040, and 1045) for the formation algorithm and run a simulation experiment n=10 times for each group. Average convergence times measured for these groups of robots along with 95\% confidence interval. We analyzed that group of 1036 robots able to form a complete ``star" shape by running n=10 simulation experiments. Robots showed average convergence time of 861.8 iterations and 95\% confidence interval ranges from 599.1 to 1124.49 iterations as shown in figure \ref{fig:star}. 

\subsubsection{Wrench shape}
\label{sub-sec:wrench}
\begin{figure}[H]
\begin{subfigure}{0.5\textwidth}
   \includegraphics[width = 6.7 cm, height = 5 cm]{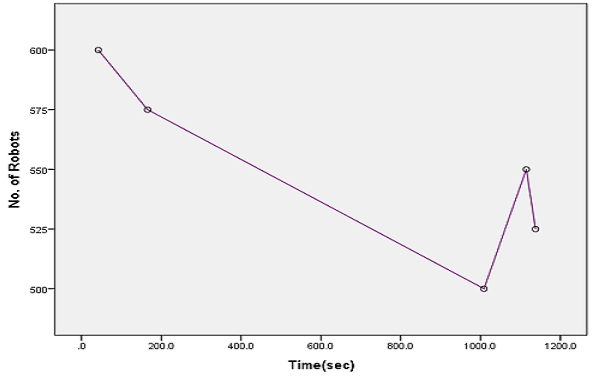}
   \caption{}
   \label{fig:wrenchConTime}
  \vspace*{0.3 cm}
\end{subfigure}%
\begin{subfigure}{0.5\textwidth}
  \includegraphics[width =6.7 cm, height = 5 cm]{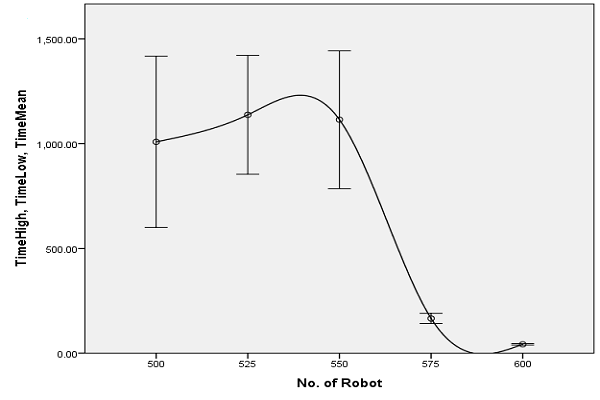}
\caption{}  
 \label{fig:wrenchConI}
    \vspace*{0.3 cm}
\end{subfigure}\\
\begin{subfigure}{0.5\textwidth}
  \includegraphics[width=8.5 cm, height = 5 cm]{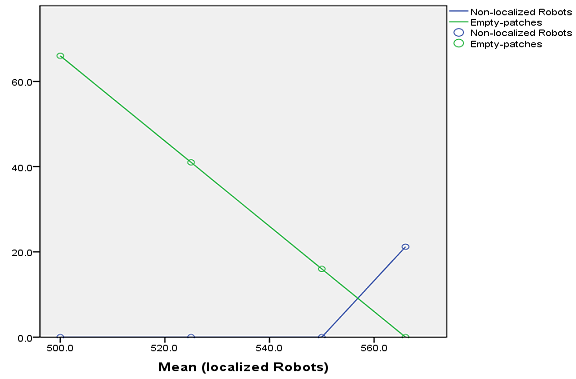}
\caption{}  
  \label{fig:wrenchShapCom}
\end{subfigure}
\begin{subfigure}{0.5\textwidth}
  \includegraphics[width=8.5 cm, height = 5 cm]{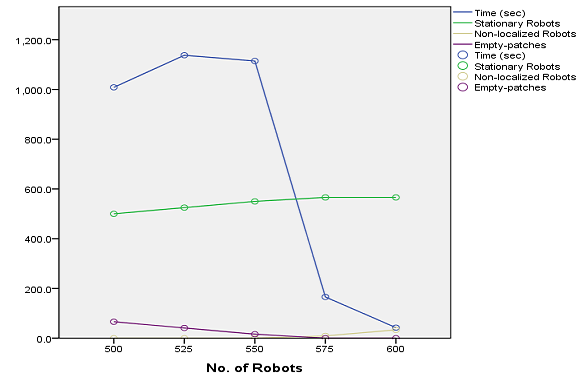}
\caption{}  
  \label{fig:wrechConVar}
\end{subfigure}
\caption{Experiment results for the ``wrench" shape, show the time convergence and shape completeness for five different robot groups.}
\label{fig:wrench}
\end{figure}

The shape information given to the algorithm after the imported image of the ``wrench" to the world environment of simulation. Algorithm applied to five groups of robots (500, 525, 550, 575, and 600) by running simulation experiments for n=10 times for every group. We calculated average convergence time for these groups of robots along with 95\% confidence interval as well as completeness of ``wrench" by these groups of robots. We observed that 566 robots made complete ``wrench" shape with average convergence time of 984.4 iterations and 95\% confidence interval ranges from 804.48 to 1164.31 iterations as shown in figure \ref{fig:wrench}.  

\subsubsection{K-letter shape}
\label{sub-sec:k}
\begin{figure}[H]
\begin{subfigure}{0.5\textwidth}
   \includegraphics[width = 6.7 cm, height = 5 cm]{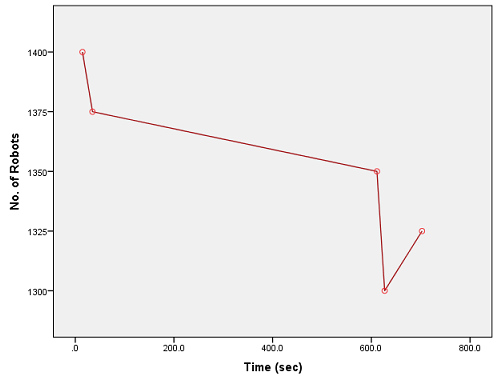}
\caption{}  
  \label{fig:kConTime}
  \vspace*{0.3 cm}
\end{subfigure}%
\begin{subfigure}{0.5\textwidth}
  \includegraphics[width =6.7 cm, height = 5 cm]{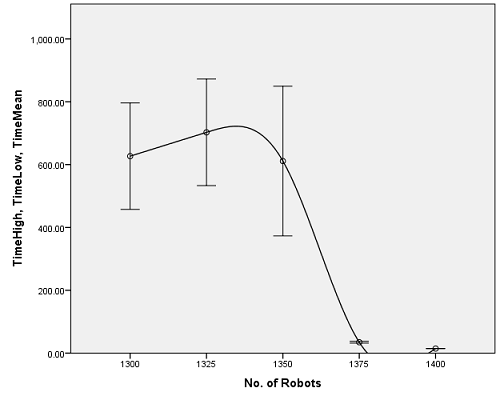}
\caption{}  
  \label{fig:kConI}
    \vspace*{0.3 cm}
\end{subfigure}\\
\begin{subfigure}{0.5\textwidth}
  \includegraphics[width=8.5 cm, height = 5 cm]{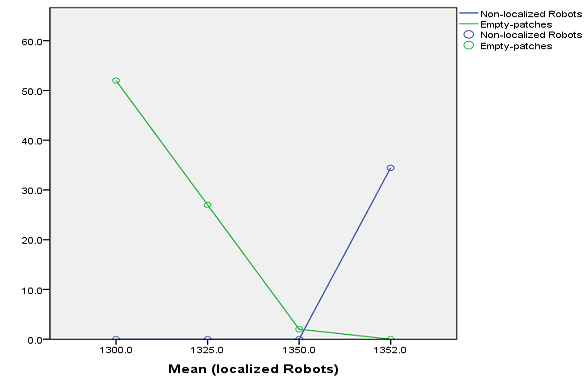}
\caption{}  
  \label{fig:kShapCom}
\end{subfigure}
\begin{subfigure}{0.5\textwidth}
  \includegraphics[width=8.5 cm, height = 5 cm]{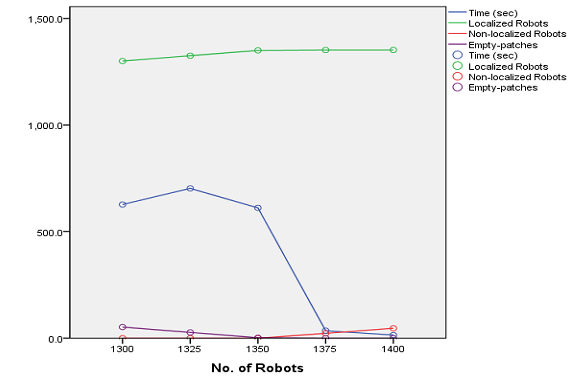}
\caption{}  
  \label{fig:kConVar}
\end{subfigure}

\caption{Experiment results for the ``k-letter" shape, show the time convergence and shape completeness for five different robot groups.}
\label{fig:k}
\end{figure}

The proposed shape formation algorithm applied to the five groups (1300, 1325, 1350, 1375, and 1400) of robots for the formation of a ``k-letter" shape. The simulation experiments executed  n=10 times for these groups to calculate average convergence time with 95\% confidence interval and shape completeness status. Figure \ref{fig:k} shows these results, where the complete shape achieved by 1352 robots group at average convergence time of 628.2 and 481.11 to 775.28 iterations for 95\% confidence interval.

\subsubsection{Rectangle shape}
\label{sub-sec:rec}
\begin{figure}[H]
\begin{subfigure}{0.5\textwidth}
   \includegraphics[width = 6.7 cm, height = 5 cm]{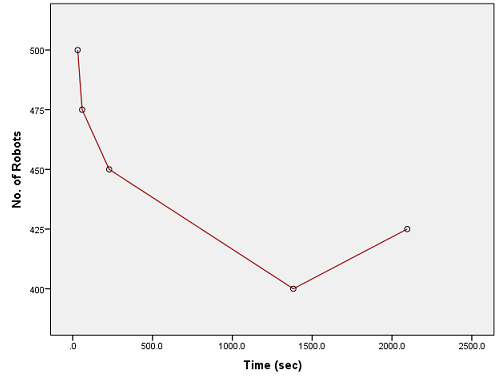}
\caption{}  
  \label{fig:recConTime}
  \vspace*{0.3 cm}
\end{subfigure}%
\begin{subfigure}{0.5\textwidth}
  \includegraphics[width =6.7 cm, height = 5 cm]{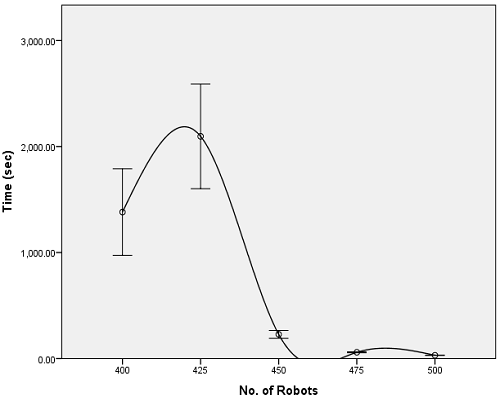}
\caption{} 
  \label{fig:recConI}
    \vspace*{0.3 cm}
\end{subfigure}\\
\begin{subfigure}{0.5\textwidth}
  \includegraphics[width=8.5 cm, height = 5 cm]{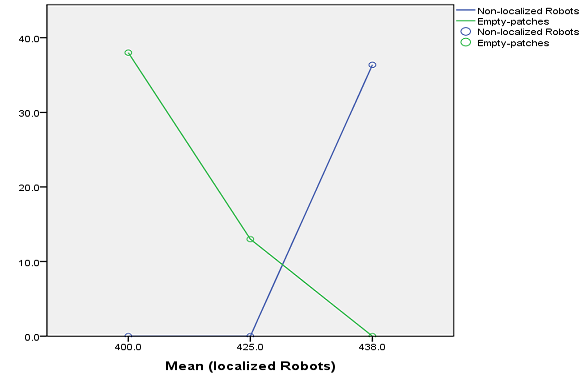}
\caption{} 
  \label{fig:recShapCom}
\end{subfigure}
\begin{subfigure}{0.5\textwidth}
  \includegraphics[width=8.5 cm, height = 5 cm]{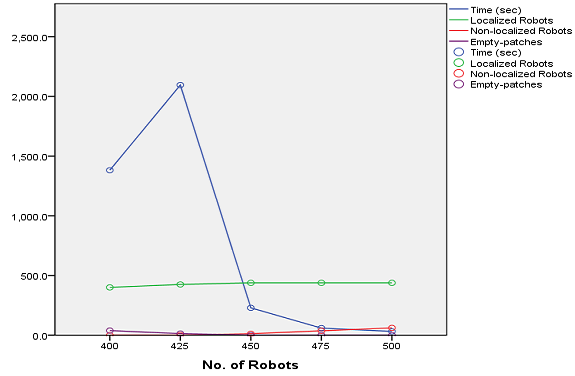}
\caption{} 
  \label{fig:recConVar}
\end{subfigure}

\caption{Experiment results for the ``rectangle" shape, show the time convergence and shape completeness for five different robot groups.}
\label{fig:rec}
\end{figure}

We performed n=10 simulation experiments upon five groups (400, 425, 450, 475, and 500) of robots for the formation of ``rectangle" shape by using proposed shape formation algorithm. We measured average convergence time with a confidence interval of 95\% and completeness of the shape  for all the groups. It can be observed from the figure \ref{fig:rec},  that the ``rectangle" shape completely formed by the 438 robots with average convergence time of 1479.8 iterations along with the 95\% confidence interval of 1253.58 to 1706.01 iterations. 

\subsubsection{Tyre shape}
\label{sub-sec:tyre}
\begin{figure}[H]
\begin{subfigure}{0.5\textwidth}
   \includegraphics[width = 6.7 cm, height = 5 cm]{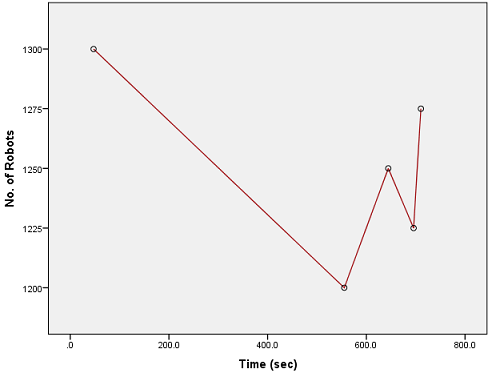}
\caption{}  
  \label{fig:tyreConTime}
  \vspace*{0.3 cm}
\end{subfigure}%
\begin{subfigure}{0.5\textwidth}
  \includegraphics[width =6.7 cm, height = 5 cm]{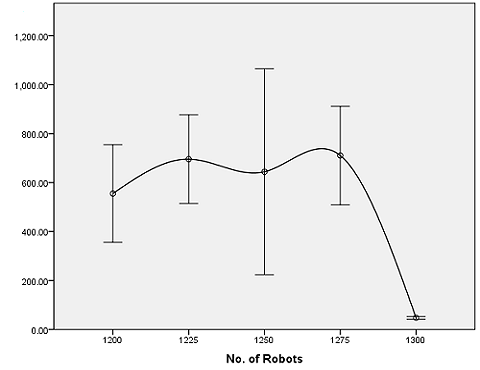}
\caption{}  
  \label{fig:tyreConI}
    \vspace*{0.3 cm}
\end{subfigure}\\
\begin{subfigure}{0.5\textwidth}
  \includegraphics[width=8.5 cm, height = 5 cm]{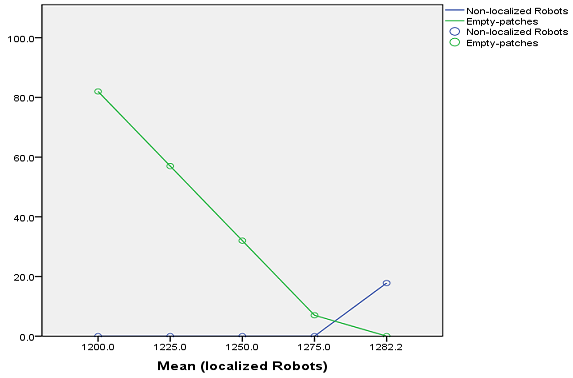}
\caption{} 
  \label{fig:tyreShapCom}
\end{subfigure}
\begin{subfigure}{0.5\textwidth}
  \includegraphics[width=8.5 cm, height = 5 cm]{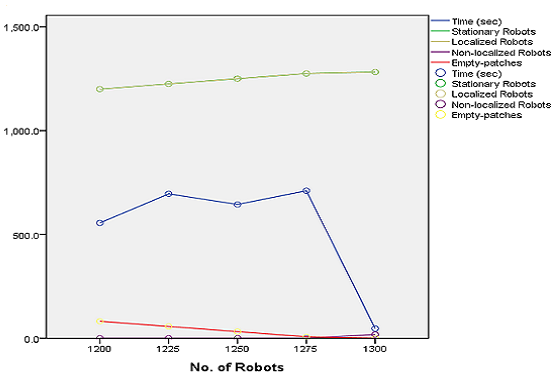}
\caption{}  
  \label{fig:tyreConVar}
\end{subfigure}

\caption{Experiment results for the ``tyre" shape, show the time convergence and shape completeness for five different robot groups.}
\label{fig:tyre}
\end{figure}

The five groups (1200, 1225, 1250, 1275, and 1300) of robots involved in n=10 simulation-based experiments for the ``tyre" shape formation using proposed algorithm. At the completion of experiments, we observed the average convergence time with 95\% confidence interval and completeness of the formed shape. The robot swarm of size 1282 formed the desired shape at average convergence time of 586.7 iterations with 95\% confidence interval value as 431.93 to 741.46 as shown in figure \ref{fig:tyre}.

\subsubsection{Spinner shape}
\label{sub-sec:spinner}
\begin{figure}[H]
\begin{subfigure}{0.5\textwidth}
   \includegraphics[width = 6.7 cm, height = 5 cm]{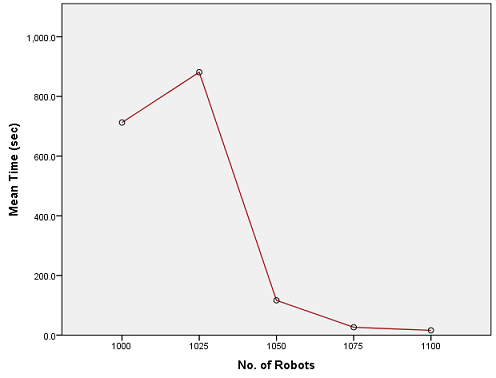}
\caption{} 
  \label{fig:spinnerConTime}
  \vspace*{0.3 cm}
\end{subfigure}%
\begin{subfigure}{0.5\textwidth}
  \includegraphics[width =6.7 cm, height = 5 cm]{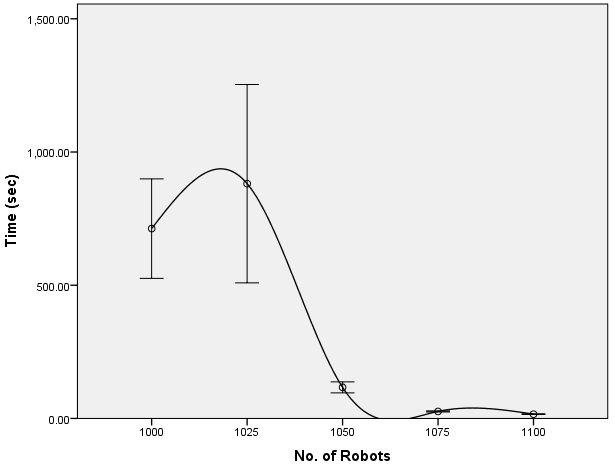}
\caption{}  
  \label{fig:spinnerConI}
    \vspace*{0.3 cm}
\end{subfigure}\\
\begin{subfigure}{0.5\textwidth}
  \includegraphics[width=8.5 cm, height = 5 cm]{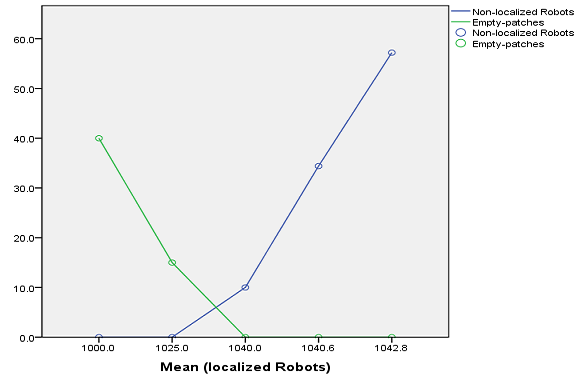}
\caption{}  
  \label{fig:spinnerShapCom}
\end{subfigure}
\begin{subfigure}{0.5\textwidth}
  \includegraphics[width=8.5 cm, height = 5 cm]{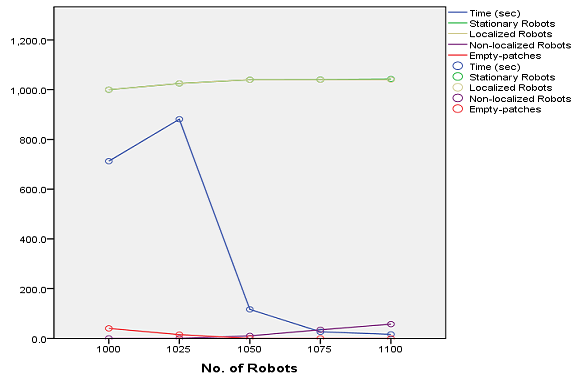}
\caption{} 
  \label{fig:spinnerConVar}
\end{subfigure}

\caption{Experiment results for the ``spinner" shape, show the time convergence and shape completeness for five different robot groups.}
\label{fig:spinner}
\end{figure}

We applied our proposed shape formation algorithm upon desired ``spinner" shape selected by the shape-type parameter for the five groups (1000, 1025, 1050, 1075, and 1100) of robots. The results achieved through n=10 number of simulation-based experiments demonstrated the average convergence times with the 95\% confidence interval and completeness of the desired ``spinner" shape. From figure \ref{fig:spinner}, it is observed that the group of 1040 robots formed a complete shape with average convergence time 802.5 iterations with  457.41 to 1147.58 confidence interval of 95\%.\\

\begin{figure}[H]
\begin{subfigure}{0.5\textwidth}
\centering
\includegraphics[width = 7.0 cm, height = 5.0 cm]{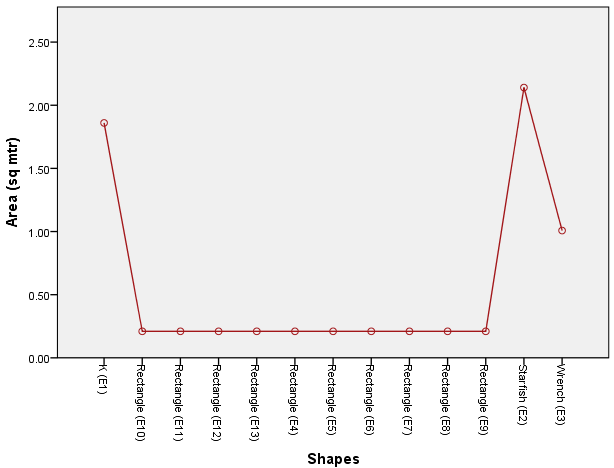}
\caption{}
\end{subfigure}
\begin{subfigure}{0.5\textwidth}
\centering
\includegraphics[width = 7.0 cm, height = 5.0 cm]{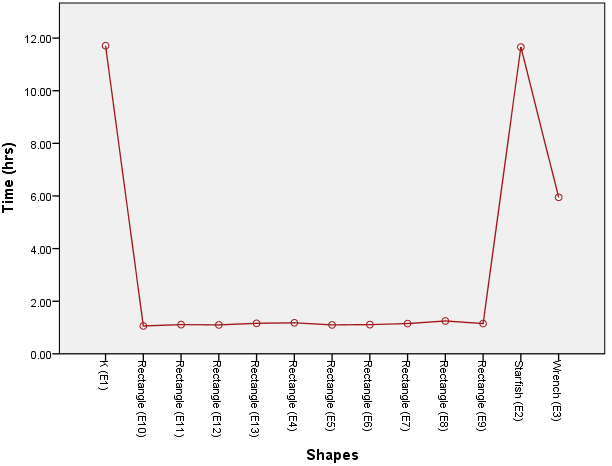}
\caption{}
\end{subfigure}
\begin{subfigure}{0.5\textwidth}
\vspace{0.5 cm}
\hspace{4.0 cm }
\includegraphics[width = 10.0 cm, height = 5.0 cm]{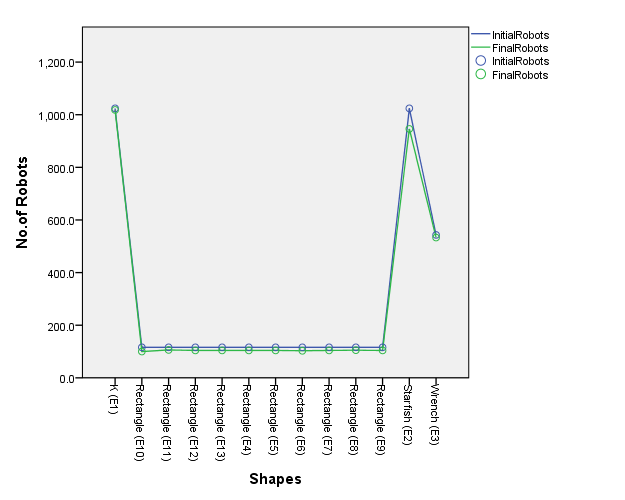}
\end{subfigure}

\caption{Replicated results for the data given in table \ref{tbl:base-results} obtained from study by Rubenstein et al.}
\label{fig:base-result}
\end{figure}

Now we discuss the results previously achieved by the Rubenstein et al. in \cite{rubenstein2014programmable} by performing thirteen-experiments using thousand programmable robots swarm in order to make four predefined shapes by the self-assembly of robots. First three experiments were for ``k-letter", ``star", and ``wrench" shapes while remaining ten experiments were for a ``rectangle" shape. The results described in term of a number of robots involved in self-assembly process for the specific shape, time for shapes completion, and dimensions of the final shapes as shown in the table \ref{tbl:base-results}. The shape completion time measured for the ``star" shape is 11.66 hours by 946 robots, the ``k-letter" consumes 11.71 hours for 1018 robots, ``wrench" formed in 5.95 hours by 533 robots, and the average time for ``rectangle" shape formation is 1.7 hours by using average 103.8 robots. We illustrate these results explicitly in the figure \ref{fig:base-result}.

\subsection{Previous studies}
\label{sec:RW}

Chen and Crilly suggested a cross-domain framework to relate applications of complex designs \cite{chen2016describing}. The author presented a comparative analysis based on the derived parameters from the three aspects which are characterization, purpose, and realization of complex systems design. The analysis depicts that the framework is capable of studying and designing complex systems across different domains. However, framework did not cover all the aspects of designing complex systems but can be used as a flexible and extendable framework by further partitioning the major aspects of complex system design.\\

An algorithm called ``potential field control"
  presented by Qin et al. in \cite{qin2013formation}, based on an exponential functions in the form of bounded artificial forces for formation control in swarm robotic systems. The simulation results showed the algorithm robust in term of swarm intelligence behavior in the real-world scenarios like local minima, avoiding obstacles and deformation of design. The proposed system more likely to generate local minima and less adaptive to the environment for the formation, based on the multi-robots.\\

Meng et al. in \cite{meng2013morphogenetic}, used gene regulatory network (GRN) in morphogenetic approach for addressing complex shapes formation. The results demonstrated the robustness and effectiveness of the proposed approach in a decentralized environment for swarm robots in order to make the complex shapes. The proposed system is dependent on the size of the swarm and localization scheme is not much efficient for the global coordinate system.\\ 

The shape formation system using potential function based scheme, which controlled the formation of multiple shapes by addressing the issues of the collision, distance sensing and local minima presented by Jung and Kim in \cite{jung2014potential}. The proposed system is scalable for the size of the swarm and showed the effectiveness of the scheme for the formation of shapes like diamond, elliptical and heart by using agents in a swarm. The system showed stability, which analyzed by using ``lyapunov" approach. While system lacks the dynamic behavior for the stochastic or obstacle-based environment, so system is unable to relate the real world scenario.\\  
 
An agent-based self-organized algorithm proposed by Niazi in \cite{niazi2014emergence}, for the modeling of emergent and dynamic snake-like structure. The achieved results showed the resilience behavior of the proposed approach, using agent-based modeling for the self-organization. The proposed algorithm lacks the decentralized environment where multiple factors may affect the agent.\\ 

MAS model based on the topology of mixed coupling proposed by Chen and Chu in \cite{chen2013multi}, for the self-organized specific pattern or geometric shapes formation. The result showed the formation convergence for symmetric coupling as uniform behavior and complex for other. While using this, the control for coordination design can be achieved. However, the system is scalable only for a limited number of agents for the proposed approach to be effective in a multi-agent system model, which made it less adaptive.\\    

In \cite{cheah2009region}, Cheah et al. proposed control method by using  ``lyapunov" like function to analyze the convergence of robots with dynamic group and positions of robots in a region. The simulation experiment showed the analysis for the convergence of robots in the multi-robot system for shape formation controller. The proposed method lacks environment adaptability and considered only the obstacle-free environment.\\  

An artificial GRN based framework presented by Guo et al. in \cite{guo2012morphogenetic} for the self-organized pattern generation and covering boundaries in the multi-robot system.  For analyzing the effectiveness of the proposed framework, various case studies were taken and results showed the adaptive behavior in term of group size and decentralized nature of the self-organized algorithm. However, the current scheme is based on an assumption about the robots accurate localization in the coordinate system, which needs to be improve in term of efficient localization scheme.\\      

In \cite{arvin2011imitation}, Arvin et al. in their work implemented aggregation of honeybee based on the various parameter values as well as proposed two variations of honeybee aggregation called ``waiting time" and ``dynamic velocity". The experiments observed variation in collective behavior of swarm robotic systems against other algorithms. The results also showed the efficiency of the proposed variation of honey bee aggregation in term of aggregation time. The proposed system ignored local communication among homogeneous robots.\\   
             
In \cite{ekanayake2010formations}, Ekanayake and Pathirana presented a scalable control algorithm for robots in a swarm by using artificial forces in order to form predefined 2D shapes. The simulation results illustrated the robustness of an algorithm. The performance analyzed by considering the issues in real applications, like communication distance and localization. However, the proposed architecture lacked the heterogeneity among swarm robots, ignored the obstacle-based environment and not scalable for making contours that are more complex.\\     

Zhang and Pathirana in \cite{zhang2011optimization}, proposed a solution to the formation control issue regarding a group of mobile robots for making random geometric patterns. A presented algorithm based on the optimization technique is distributed and focused on the desired geometric pattern for robot current arrangement. The strategy for movement assured continuous motion until the formation of desired geometric pattern. The system achieved the process of optimization which depends upon the perfect positioning for the desired pattern with respect to the current configuration of robots and communication that best suited for robots in the group. While system lacks the local interaction among robots and for a small set of robots, it optimizes independently.\\
          
The positioning algorithms are inspired by the systems having the model matching capability while traditional model matching systems are not much distributed. An algorithm presented by Seckin et al. in \cite{secckin2016feature}, designed to work in a distributed manner comprises of two stages, offline and online. Data collection and computation of the positions is completed in the offline stage, while transfer and communication of data in another stage. The results illustrated that the positions of the robots are based on their features given in the input image. The comparison made in the study is based only on the optimality for the various feature-matching algorithms.\\

The self-assembly based shape formation for swarm robotic system presented by Rubenstein et al. in \cite{rubenstein2014programmable}. An experiment carried out for the formation of shapes like ``starfish", ``wrench", ``K-letter", and ``rectangle" by applying collective behavior algorithm. However, at the start of the self-assembly process, robots are placed in aggregated form with fix locations and no random localization. Movement for robots is in sequential edge following manner without random motion. Self-assembly process depends on swarm size for formation, therefore no scalability for swarm size. The system limited only to solid shapes with no holes, this unable to form a complex symmetries. Furthermore, it also ignored the decentralized environment. \\

The self-organization in the swarm robotic system presented by Copenhagen et al. in \cite{copenhagen2016self} for sorting limits of the heterogeneous robots using multi-agents, made clusters based on the alignments of agents. There is an aggregation and segregation of agents in a swarm to make clusters. The proposed system has swarm size limitation, as well as clusters, comprise of both aligned and non-aligned agents instead of having separate clusters of each agent group in a two-dimensional space. The system also lacks the ABM approach for self-organized sorting of swarm robots.\\

We performed comparative analysis by reviewing literature discussed in this section, on the bases of the techniques used along with with the challenges faced or constraints in the swarm robotic systems for the self-organized formation task. The findings and comparisons of the previous work expressed in tabular form given in the table \ref{tbl:related-work}.
\section{Conclusion}
\label{sec:conclusion}

In this paper, we presented simulation model using ABM for the self-organized shape formation task in swarm robotic system. We also illustrated the formal descriptions for our model to validate all the informal requirements of the system. In this context, we proposed a self-organized shape formation algorithm, having no centralized controller instead rely on the individual actions and behaviors of agents for making flexible shapes. We performed number of simulation-based experiments for six different shapes and observed the convergence of agents for making these shapes. Simulation results showed the effectiveness of the proposed approach in term of agents adaptive behavior in the large-scale swarm for a decentralized environment, by successfully achieving the formation of six diverse shapes including hole-based shape. The behavior analysis for our proposed approach demonstrated that for a larger number of robots, the average time convergence is less than that for smaller swarm size in self-organization process of shape formation. In Addition to the future work, we intend to build the same scenario of self-organized flexible shapes formation in three-dimensional space with more adaptability feature in swarm robots.\\ 

\newpage


\begin{table}[H]
\centering
\caption{Results achieved by Rubenstein et al. in  \cite{rubenstein2014programmable}.}
\label{tbl:base-results}
\begin{adjustbox}{width=0.7\textwidth}
\small

\begin{tabular}{c c c c c}
\hline
\textbf{Desire shape} & \textbf{\makecell{Robots at\\ start}} &\textbf{\makecell{Robots in\\ final shape}} & \textbf{\makecell{Completion time\\ (hrs)}} & \textbf{\makecell{Shape dimension\\ (m)}}\\ \hline
K(E1) & 1024 &1018 & 11.71 & 1.39 x 1.34\\ 
Starfish(E2) & 1024 & 946 &	11.66 &	1.47 x 1.47 \\ 
Wrench(E3) & 543 & 533	&5.95 &	 1.60 x 0.63  \\ 
Rectangle(E4) & 116  & 104&	1.18	&0.35 x 0.61 \\  
Rectangle(E5) & 116 & 104&	1.10&	0.35 x 0.61 \\ 
Rectangle(E6) & 116 & 103	&1.11&	0.35 x 0.61  \\ 
Rectangle(E7) & 116 & 104&	1.15&	0.35 x 0.61  \\ 
Rectangle(E8) & 116 & 105	&1.25&	0.35 x 0.61  \\ 
Rectangle(E9) & 116 & 104	&1.15&	0.35 x 0.61  \\ 
Rectangle(E10) &116 & 100&	1.06&	0.35 x 0.61 \\ 
Rectangle(E11) &116 & 106&	1.11&	0.35 x 0.61 \\ 
Rectangle(E12) &116 & 104&	1.10&	0.35 x 0.61 \\
Rectangle(E13) &116 & 104&	1.16&	0.35 x 0.61  \\ 
\hline

\end{tabular}
\end{adjustbox}
\end{table}


\begin{table}[H]
\begin{center}
\caption{Experiments summary for proposed shape formation algorithm for all desired shapes.}\label{tbl:new-results}
\small
\begin{tabular}{c c c c c}
\hline
\textbf{Shape} & \textbf{Dimensions} &\textbf{ No. of robots } & \textbf{\makecell{Avg. time\\(sec)}} & \textbf{\makecell{Confidence\\ interval (95\%)}} \\ \hline
Star & 464 x 438 &1036 & 861.8 & 599.1 --- 1124.49 \\ 
K-letter & 446 x 462 & 1352 & 628.2 & 481.11 --- 775.28 \\ 
Wrench & 466 x 430 & 566 & 984.4 & 804.48 --- 1164.31  \\ 
Rectangle & 409 x 352  & 438 & 1479.8 &	1253.58 --- 1706.01 \\  
Tyre & 472 x 413 & 1282 &	586.7  & 431.93 --- 741.46\\ 
Spinner & 458 x 427 & 1040 & 802.5 & 457.41 --- 1147.58\\ \hline

\end{tabular}
\end{center}
\end{table}


\begin{table}[H]
\centering
\caption{Comparative analysis of previous studies.}
\label{tbl:related-work}
\begin{adjustbox}{width=1\textwidth}
\begin{tabular}{c c c c c c c c c c}

\hline
\textbf{Ref.} & 
\textbf{Technique} &\textbf{Real exp.} & \textbf{ABM} & \textbf{\makecell{Swarm \\ limit}}  & \textbf{\makecell{Dynamic \\localization}} & \textbf{\makecell{Random \\motion}}  & \textbf{\makecell{Complex \\shapes}} & \textbf{\makecell{Obstacle-\\avoidance}} & \textbf{\makecell{Heterogeneous \\swarm}}\\ \hline

\cite{qin2013formation} & 
 Potential-field control algorithm & \xmark	 
 &  \xmark & \xmark  & \cmark & \cmark & \xmark & \cmark & \xmark\\ \hline
 
\cite{meng2013morphogenetic} & 
GRN based morphogenetic approach & \cmark 
 &  \xmark & \cmark & \xmark & \cmark & \cmark & \cmark & \xmark  \\ \hline
\cite{jung2014potential} & 
Potential function with  lyapunov approach &\xmark 
&  \xmark & \xmark & \cmark & \cmark & \xmark & \xmark & \xmark  \\ \hline 
\cite{niazi2014emergence} & 
Self-organization algorithm &\xmark 
&  \cmark & \xmark & \cmark & \cmark & \cmark & \xmark & \xmark 		 \\ \hline
\cite{chen2013multi} & 
Mix coupling topology	& \xmark 
&  \xmark & \cmark & \cmark & \cmark & \cmark & \cmark & \cmark 	  \\ \hline
\cite{cheah2009region} & 
Lyapunov  function & \xmark 
&  \xmark & \xmark & \cmark & \cmark & \xmark & \xmark & \xmark \\ \hline
\cite{guo2012morphogenetic} & 
GRN based morphogenetic approach 	& \xmark  
& \xmark & \xmark  & \xmark & \cmark & \xmark & \cmark & \xmark \\ \hline
 \cite{arvin2011imitation} & 
Honey-bee aggregation & \cmark 
& \xmark & \cmark & \cmark & \cmark & \xmark & \cmark & \xmark \\ \hline
\cite{ekanayake2010formations} & 
Control algorithm & \xmark 
& \xmark & \cmark & \cmark & \cmark & \xmark & \xmark & \xmark \\ \hline
\cite{zhang2011optimization} & 
Optimization based formation algorithm & \xmark 
& \xmark & \cmark & \cmark & \cmark & \xmark & \xmark & \xmark \\ \hline
\cite{secckin2016feature} & 
Positioning algorithm & \xmark 
& \xmark & \cmark & \xmark & \xmark & \xmark & \xmark & \xmark \\ \hline
\cite{rubenstein2014programmable} & 
Self-assembly algorithm & \cmark 
& \xmark & \cmark & \xmark & \xmark & \xmark & \xmark & \cmark \\ \hline
\cite{copenhagen2016self} & 
Self-organized sorting & \xmark 
& \xmark & \cmark & \cmark & \cmark & \xmark & \cmark & \cmark \\ \hline

\end{tabular}
\end{adjustbox}
\end{table}



\newpage

\begin{center}
\begin{longtable}[c]{ c  p {15 cm}} 
 
 \textbf{Ref.} \textbf{:} & \textbf{Caption}  \\ 
 \hline
 \endfirsthead
 
\textbf{Ref.} \textbf{:} & \textbf{Caption}  \\ 
 \hline
 \endhead
 
 \hline
 \endfoot 
 
 \endlastfoot
 
\\Fig \ref{fig:base-result}: & \textbf{a)} It represents the dimension or area in meters for the final formed shapes, \textbf{b)} The shape completion time measured in hours, shown for each of the thirteen experiments, \textbf{c)} A number of robots initially involved and a final number of robots after shape completion.\\ \\\hline
 
\\Fig \ref{fig:model}: & It shows the design interface for our agent-based model for the self-organized shape formation. The image shows the GUI of an environment world, where agents (purple colored) as robots placed at random locations while seed robots (red colored) placed near the origin at center of the world. There are sliders for controlling a number of robots and vision-radius for communication among neighbors. A chooser for selecting the desired shape, buttons for initializing and running simulation. The monitors are for monitoring the states of robots and plot for the robot convergence rate, which changes during simulation. \\ \\\hline

\\Fig \ref{fig:allShapes}: & Final formed shapes by swarm robots: \textbf{(a)} A complete ``star" shape formed by 1036 robots including four seed robots in our simulation model, \textbf{(b)} A final ``wrench" shape formed through simulation in our model for self-organized shape formation by the convergence of 566 robots initially placed and move randomly in the world, \textbf{(c)} A complete ``k-letter" shape formed by 1352 robots including four seed robots in our simulation model, \textbf{(d)} A ``rectangle" shape formed through simulation in our model for self-organized shape formed by the convergence of 438 robots, \textbf{(e)} A ``tyre" shape formed through simulation in our model for self-organized shape formation by the convergence of 1282 robots initially placed and move randomly in the world,  \textbf{(f)} A final ``spinner" shape formed through simulation in our model for self-organized shape formation by the convergence of 1040 robots.\\  \\\hline

\\Fig \ref{fig:all-result}: & The experiment results for all the desired shapes: \textbf{(a)} It shows the convergence rate for all six desired shapes, i.e. the total number of robots involved in forming the desired shapes with respect to the time taken for the formation, \textbf{(b)} It represents the convergence time with 95\% confidence interval for all six shapes. 
\\ \\\hline

Fig \ref{fig:star}: & The simulation results for the ``star" shape formation achieved by performing a number of experiments as n=10 for five (1025, 1030, 1035, 1040, and 1045) groups of robots and averaging their results: \textbf{(a)} The average convergence time for a different number of robots ranges from 125.7 to 739.7 iterations, shows less time for large robot group size as compare to small group size, \textbf{(b)} It shows the 95\% of the confidence interval for convergence time for the above-mentioned groups as 105.96 to 1044.26 iterations, \textbf{(c)} The shape completeness shows the average number of localized and stationary robots as 1025 to 1036 robots with the remaining unlocalized robots 0 to 8.9 and empty spaces or patches within the shape range from 0 to 11 for all groups respectively. It showed that the group size of 1036 robots completely formed the desired ``star" shape without leaving any empty patches within the shape and unlocalized extra robots. Below this group size, there will be empty spaces and shape will be incomplete while above this group size, there are unlocalize robots still remaining and waiting to become localize and stationary. Thus, they separate themselves from the self-organization process, \textbf{(d)} In this graph, convergence behavior variations are shown for the ``star" shape. \\ \\\hline

\\Fig \ref{fig:wrench}: & The ``wrench" shape formation achieved by performing n=10 simulation experiments for five (500, 525, 550, 575, and 600) groups of robots and averaging their results: \textbf{(a)} The average convergence time for a different number of robots ranges from 42.3 to 1137.7 iterations, shows less time for the large robots group than small group size, \textbf{(b)} It shows the convergence time for the above-mentioned groups of robots with 95\% of the confidence interval as 38.81 to 1443.49 iterations, \textbf{(c)} The ``wrench" shape completeness measures as the average number of localized and stationary robots which are 500 to 566 with the remaining unlocalized are 0 to 33.4 and empty patches within the shape are 0 to 66 for each of the robot group size respectively. The group size of 566 robots completely formed the desired ``wrench" shape without leaving any empty patches within the shape and unlocalize extra robots. Below this group, the shape will be incomplete and above this size, there will more unlocalize robots outside the shape, \textbf{(d)} The variations in convergence in ``wrench" shape comprises average time convergence, shape completeness for each of the robots group sizes. \\\\\hline

\\Fig \ref{fig:k}: & The simulation results for the ``k-letter" formation by performing n=10 for five (1300, 1325, 1350, 1375, and 1400) groups of robots:  \textbf{(a)} The average convergence time lies between 14.6 to 702.7 iterations for all groups,  \textbf{(b)} 95\% of the confidence interval 14.1 to 872.34 of convergence time for all groups,  \textbf{(c)} The completeness shows the average number of localized and stationary robots ranges from 1300 to 1352 and 0 to 46.4 remaining unlocalize robots. The empty-patches within the shape are 0 to 52 for all groups. The group of 1352 robots completely formed a ``k-letter" shape without leaving any empty-patches and unlocalize extra robots, \textbf{(d)} The convergence variations for the desired ``k-letter" shape are shown for each of robots group. \\ \\\hline

Fig \ref{fig:rec}: & The results of (n=10) simulation experiments for ``rectangle" shape formation by five (400, 425, 450, 475, and 500) groups of robots: \textbf{(a)} The value lie between 31.2 to 2095.6 iterations for the average convergence time of all five groups, it shows less time for the large robots group than small group size, \textbf{(a)} The 95\% confidence interval value as 29.39 to 2589.03 for the convergence times of all groups, \textbf{(c)} The completeness of the ``rectangle" shape depicts average number of localized robots as 400 to 438, unlocalize robots 0 to 60.8, and 0 to 38 empty-patches for all groups. It observed that the group of 438 robots completely formed ``rectangle" shape without leaving any empty-patches and unlocalize robots. While below this group, the incomplete shape formed and unlocalize robots remained above this group size, so they separated themselves from the formation process, \textbf{(d)} The convergence variations are shown as average time convergence and shape completeness for each group of robots.\\ \\\hline

\\Fig \ref{fig:tyre}: & The five groups (1200, 1225, 1250, 1275, and 1300)of robots used in simulation experiment for the ``tyre" shape and the obtained results are averaged : \textbf{(a)} The convergence time for a different number of robots lie between 47.2 to 710.5 iterations, \textbf{(b)} It shows the convergence time with 95\% confidence interval as 41.82 to 1065.37 for all groups, \textbf{(c)} The shape completeness shows the average number of localized and stationary as 1200 to 1282 robots while remaining unlocalize are 0 to 17.8 robots with 0 to 82  value for the number of empty-spaces within the shape. A group of 1282 robots completely formed the desired ``tyre" shape without leaving any empty patches within the shape, \textbf{(d)}  The average time convergence and shape completeness for each group of robots is shown together as a variation in convergence .\\ \\\hline

\\Fig \ref{fig:spinner}: & The simulation results for the ``spinner" shape formation by performing n=10 experiments for five groups (1000, 1025, 1050, 1075, and 1100) of robots: \textbf{(a).} The average convergence time is between 16.1 to 881.2 iterations, it showed less time for the large group than small number of robots, \textbf{(b)} The convergence time with 95\% confidence interval of 15.57 to 1253.37 for number of robots, \textbf{(c)} The shape completeness depicts the average number of localized and stationary robots as 1000 to 1042.8 with remaining unlocalize robots as 0 to 57.2 along with 0 to 40 empty-patches within the shape for all groups. The group of 1040 robots completely formed ``spinner" shape without leaving any empty patches and unlocalize robots. While less than this formed  the incomplete shape and above this group size, there were unlocalize robots left, \textbf{(d)} Convergence variations of the ``spinner" showing average time convergence and shape completeness for each of robots group. \\ 

\end{longtable}
\end{center}

\newpage
\textbf{References}
\bibliography{sample}

\begin{thebibliography}{10}

\bibitem{oh2017bio}
Hyondong Oh, Ataollah~Ramezan Shirazi, Chaoli Sun, and Yaochu Jin.
\newblock Bio-inspired self-organising multi-robot pattern formation: A review.
\newblock {\em Robotics and Autonomous Systems}, 91:83--100, 2017.

\bibitem{schmickl2008trophallaxis}
Thomas Schmickl and Karl Crailsheim.
\newblock Trophallaxis within a robotic swarm: bio-inspired communication among
  robots in a swarm.
\newblock {\em Autonomous Robots}, 25(1):171--188, 2008.

\bibitem{haghighi2012multi}
Reza Haghighi and Chien-Chern Cheah.
\newblock Multi-group coordination control for robot swarms.
\newblock {\em Automatica}, 48(10):2526--2534, 2012.

\bibitem{kumar2010segregation}
Manish Kumar, Devendra~P Garg, and Vijay Kumar.
\newblock Segregation of heterogeneous units in a swarm of robotic agents.
\newblock {\em IEEE Transactions on Automatic Control}, 55(3):743--748, 2010.

\bibitem{de2016distributed}
Alan~Oliveira de~S{\'a}, Nadia Nedjah, and Luiza de~Macedo~Mourelle.
\newblock Distributed efficient localization in swarm robotic systems using
  swarm intelligence algorithms.
\newblock {\em Neurocomputing}, 172:322--336, 2016.

\bibitem{cheng2005robust}
Jimming Cheng, Winston Cheng, and Radhika Nagpal.
\newblock Robust and self-repairing formation control for swarms of mobile
  agents.
\newblock In {\em AAAI}, volume~5, 2005.

\bibitem{rubenstein2014programmable}
Michael Rubenstein, Alejandro Cornejo, and Radhika Nagpal.
\newblock Programmable self-assembly in a thousand-robot swarm.
\newblock {\em Science}, 345(6198):795--799, 2014.

\bibitem{niazi2017technical}
Muaz~A Niazi.
\newblock Technical problems with" programmable self-assembly in a
  thousand-robot swarm".
\newblock {\em arXiv preprint arXiv:1708.03341}, 2017.

\bibitem{copenhagen2016self}
Katherine Copenhagen, David~A Quint, and Ajay Gopinathan.
\newblock Self-organized sorting limits behavioral variability in swarms.
\newblock {\em Scientific reports}, 6, 2016.

\bibitem{2012cognitive}
Muaz~A Niazi and Amir Hussain.
\newblock {\em Cognitive agent-based computing-I: a unified framework for
  modeling complex adaptive systems using agent-based \& complex network-based
  methods}.
\newblock Springer Science \& Business Media, 2012.

\bibitem{karaboga2014comprehensive}
Dervis Karaboga, Beyza Gorkemli, Celal Ozturk, and Nurhan Karaboga.
\newblock A comprehensive survey: artificial bee colony (abc) algorithm and
  applications.
\newblock {\em Artificial Intelligence Review}, 42(1):21--57, 2014.

\bibitem{tan2013research}
Ying Tan and Zhong-yang Zheng.
\newblock Research advance in swarm robotics.
\newblock {\em Defence Technology}, 9(1):18--39, 2013.

\bibitem{barca2013swarm}
Jan~Carlo Barca and Y~Ahmet Sekercioglu.
\newblock Swarm robotics reviewed.
\newblock {\em Robotica}, 31(3):345--359, 2013.

\bibitem{nishikawa2016coordination}
Naoki Nishikawa, Reiji Suzuki, and Takaya Arita.
\newblock Coordination control design of heterogeneous swarm robots by means of
  task-oriented optimization.
\newblock {\em Artificial Life and Robotics}, 21(1):57--68, 2016.

\bibitem{varghese2009review}
Blesson Varghese and Gerard McKee.
\newblock A review and implementation of swarm pattern formation and
  transformation models.
\newblock {\em International Journal of Intelligent Computing and Cybernetics},
  2(4):786--817, 2009.

\bibitem{wu2016novel}
Zongsheng Wu, Weiping Fu, Ru~Xue, and Wen Wang.
\newblock A novel global path planning method for mobile robots based on
  teaching-learning-based optimization.
\newblock {\em Information}, 7(3):39, 2016.

\bibitem{bayindir2016review}
Levent Bay{\i}nd{\i}r.
\newblock A review of swarm robotics tasks.
\newblock {\em Neurocomputing}, 172:292--321, 2016.

\bibitem{niazi2008self}
Muaz~A Niazi.
\newblock Self-organized customized content delivery architecture for ambient
  assisted environments.
\newblock In {\em Proceedings of the third international workshop on Use of
  P2P, grid and agents for the development of content networks}, pages 45--54.
  ACM, 2008.

\bibitem{zhang2014swarm}
Zhongshan Zhang, Keping Long, Jianping Wang, and Falko Dressler.
\newblock On swarm intelligence inspired self-organized networking: its bionic
  mechanisms, designing principles and optimization approaches.
\newblock {\em IEEE Communications Surveys \& Tutorials}, 16(1):513--537, 2014.

\bibitem{o2010self}
Rehan O’Grady, Roderich Gro{\ss}, Anders~Lyhne Christensen, and Marco Dorigo.
\newblock Self-assembly strategies in a group of autonomous mobile robots.
\newblock {\em Autonomous Robots}, 28(4):439--455, 2010.

\bibitem{vardy2016aggregation}
Andrew Vardy.
\newblock Aggregation in robot swarms using odometry.
\newblock {\em Artificial Life and Robotics}, 21(4):443--450, 2016.

\bibitem{arvin2016investigation}
Farshad Arvin, Ali~Emre Turgut, Tom{\'a}{\v{s}} Krajn{\'\i}k, and Shigang Yue.
\newblock Investigation of cue-based aggregation in static and dynamic
  environments with a mobile robot swarm.
\newblock {\em Adaptive Behavior}, 24(2):102--118, 2016.

\bibitem{sperati2011self}
Valerio Sperati, Vito Trianni, and Stefano Nolfi.
\newblock Self-organised path formation in a swarm of robots.
\newblock {\em Swarm Intelligence}, 5(2):97--119, 2011.

\bibitem{nouyan2008path}
Shervin Nouyan, Alexandre Campo, and Marco Dorigo.
\newblock Path formation in a robot swarm.
\newblock {\em Swarm Intelligence}, 2(1):1--23, 2008.

\bibitem{bahceci2003review}
Erkin Bahceci, Onur Soysal, and Erol Sahin.
\newblock A review: Pattern formation and adaptation in multi-robot systems.
\newblock {\em Robotics Institute, Carnegie Mellon University, Pittsburgh, PA,
  Tech. Rep. CMU-RI-TR-03-43}, 2003.

\bibitem{niazi2011novel}
Muaz~A Niazi and Amir Hussain.
\newblock A novel agent-based simulation framework for sensing in complex
  adaptive environments.
\newblock {\em IEEE Sensors Journal}, 11(2):404--412, 2011.

\bibitem{niazi2009agent}
Muaz Niazi and Amir Hussain.
\newblock Agent-based tools for modeling and simulation of self-organization in
  peer-to-peer, ad hoc, and other complex networks.
\newblock {\em IEEE Communications Magazine}, 47(3), 2009.

\bibitem{niazi2011towards}
Muaz Ahmed~Khan Niazi.
\newblock {\em Towards A Novel Unified Framework for Developing Formal, Network
  and Validated Agent-Based Simulation Models of Complex Adaptive Systems}.
\newblock PhD thesis, University of Stirling, 2011.

\bibitem{laghari2016modeling}
Samreen Laghari and Muaz~A Niazi.
\newblock Modeling the internet of things, self-organizing and other complex
  adaptive communication networks: a cognitive agent-based computing approach.
\newblock {\em PloS one}, 11(1):e0146760, 2016.

\bibitem{woodcock1996using}
Jim Woodcock and Jim Davies.
\newblock {\em Using Z: specification, refinement, and proof}, volume~39.
\newblock Prentice Hall Englewood Cliffs, 1996.

\bibitem{bowen1996formal}
Jonathan~Peter Bowen.
\newblock {\em Formal specification and documentation using Z: A case study
  approach}, volume~66.
\newblock International Thomson Computer Press London, 1996.

\bibitem{d2004understanding}
Mark d'Inverno, Michael Luck, and Michael~M Luck.
\newblock {\em Understanding agent systems}.
\newblock Springer Science \& Business Media, 2004.

\bibitem{wilensky1999netlogo}
Uri Wilensky and I~Evanston.
\newblock Netlogo: Center for connected learning and computer-based modeling.
\newblock {\em Northwestern University, Evanston, IL}, 4952, 1999.

\bibitem{chen2016describing}
Chih-Chun Chen and Nathan Crilly.
\newblock Describing complex design practices with a cross-domain framework:
  learning from synthetic biology and swarm robotics.
\newblock {\em Research in Engineering Design}, 27(3):291--305, 2016.

\bibitem{qin2013formation}
Long Qin, Yabing Zha, Quanjun Yin, and Yong Peng.
\newblock Formation control of robotic swarm using bounded artificial forces.
\newblock {\em The Scientific World Journal}, 2013, 2013.

\bibitem{meng2013morphogenetic}
Yan Meng, Hongliang Guo, and Yaochu Jin.
\newblock A morphogenetic approach to flexible and robust shape formation for
  swarm robotic systems.
\newblock {\em Robotics and Autonomous Systems}, 61(1):25--38, 2013.

\bibitem{jung2014potential}
Hahmin Jung and Dong~Hun Kim.
\newblock Potential-function-based shape formation in swarm simulation.
\newblock {\em International Journal of Control, Automation and Systems},
  12(2):442--449, 2014.

\bibitem{niazi2014emergence}
Muaz~A Niazi.
\newblock Emergence of a snake-like structure in mobile distributed agents: an
  exploratory agent-based modeling approach.
\newblock {\em The Scientific World Journal}, 2014.

\bibitem{chen2013multi}
Zhifu Chen and Tianguang Chu.
\newblock Multi-agent system model with mixed coupling topologies for pattern
  formation and formation splitting.
\newblock {\em Mathematical and Computer Modelling of Dynamical Systems},
  19(4):388--400, 2013.

\bibitem{cheah2009region}
Chien~Chern Cheah, Saing~Paul Hou, and Jean Jacques~E Slotine.
\newblock Region-based shape control for a swarm of robots.
\newblock {\em Automatica}, 45(10):2406--2411, 2009.

\bibitem{guo2012morphogenetic}
Hongliang Guo, Yaochu Jin, and Yan Meng.
\newblock A morphogenetic framework for self-organized multirobot pattern
  formation and boundary coverage.
\newblock {\em ACM Transactions on Autonomous and Adaptive Systems (TAAS)},
  7(1):15, 2012.

\bibitem{arvin2011imitation}
Farshad Arvin, Khairulmizam Samsudin, Abdul~Rahman Ramli, and Masoud Bekravi.
\newblock Imitation of honeybee aggregation with collective behavior of swarm
  robots.
\newblock {\em International Journal of Computational Intelligence Systems},
  4(4):739--748, 2011.

\bibitem{ekanayake2010formations}
Samitha~W Ekanayake and Pubudu~N Pathirana.
\newblock Formations of robotic swarm: an artificial force based approach.
\newblock {\em International Journal of Advanced Robotic Systems}, 7(3):23,
  2010.

\bibitem{zhang2011optimization}
Huan Zhang and Pubudu~N Pathirana.
\newblock Optimization-based formation of autonomous mobile robots.
\newblock {\em Robotica}, 29(4):515--525, 2011.

\bibitem{secckin2016feature}
Ahmet~{\c{C}}a{\u{g}}da{\c{s}} Se{\c{c}}kin, Ceyhun Karpuz, and Ahmet {\"O}zek.
\newblock Feature matching based positioning algorithm for swarm robotics.
\newblock {\em Computers \& Electrical Engineering}, 2016.

\end{thebibliography}
\bibliographystyle{unsrt}

\end{document}